\documentclass{article}
\usepackage{ijcai22}

\usepackage{times}
\usepackage{soul}
\usepackage{url}
\usepackage[hidelinks]{hyperref}
\usepackage[utf8]{inputenc}
\usepackage[small]{caption}
\usepackage{graphicx}
\usepackage{amsmath}
\usepackage{amsthm}
\usepackage{booktabs}
\usepackage{algorithm}
\usepackage{algorithmic}
\usepackage{hyperref}
\usepackage{bm}

\usepackage{epsfig}
\usepackage{amssymb}
\usepackage{subfigure}
\usepackage{caption}
\usepackage{subfloat}
\usepackage{float}
\usepackage{enumerate}
\usepackage{multirow}
\usepackage[table,xcdraw]{xcolor}
\usepackage{makecell}
\usepackage{verbatim}
\usepackage{stfloats}

\title{Improving Robustness of Convolutional Neural Networks Using Element-Wise Activation Scaling}

\begin{document}

\maketitle

\begin{abstract}

Recent works reveal that re-calibrating the intermediate activation of adversarial examples can improve the adversarial robustness of a CNN model. The state of the arts  \cite{bai2021improving} and \cite{cifs} explores this feature at the channel level, i.e. the activation of a channel is uniformly scaled by a factor. In this paper, we investigate the intermediate activation manipulation at a more fine-grained level. Instead of uniformly scaling the activation, we individually adjust 
each element within an activation and thus propose Element-Wise Activation Scaling, dubbed EWAS, to improve CNNs' adversarial robustness. Experimental results on ResNet-18 and WideResNet with CIFAR10 and SVHN show that EWAS significantly improves the robustness accuracy. Especially for ResNet18 on CIFAR10, EWAS increases the adversarial accuracy by 37.65\% to 82.35\% against C\&W attack. EWAS is simple yet very effective in terms of improving robustness. The codes are anonymously available at \url{https://anonymous.4open.science/r/EWAS-DD64}. 

\end{abstract}

\section{Introduction}
\label{sec:intro}

Convolutional neural networks (CNNs) have demonstrated its superiority in various applications, especially for computer vision tasks, like classification, object detection and segmentation \cite{alexnet,VIT}. 
However, CNNs are found to be vulnerable to adversarial samples that are perturbed by unperceptive noises \cite{fgsm}. Adversarial attacks significantly undermine the model's robustness and threat the applicability of CNNs to some safety-critical and security-critical contexts, e.g. self-driving, person identification, etc. 
A plenty of efforts have been made to improve CNNs' adversarial robustness and these efforts can be generally divided into two categories: adversarial attacks and adversarial defence. 
Various methods are proposed to generate diverse adversarial samples 
\cite{fgsm,c&w,PGD,deepfool,pixelattack,advgan,jandial2019advgan,croce2020reliable}.

On the other hand, many works aim to defend the adversarial attacks. A number of defense methods have been proposed, such as defensive distillation \cite{defensedistillation,ard}, feature denoising \cite{xie2019feature,liao2018defense}, GAN \cite{liu2019rob}, model compression \cite{madaan2020adversarial,ye2019adversarial,gui2019model}, authentication defense \cite{chen2019deepinspect}, and adversarial training (AT) and its variants \cite{PGD,TRADES,mart,wong2020fast}. 
Recently, some works investigate the difference between natural models and AT-trained counterparts in terms of intermediate activation and propose to adjust intermediate activation for better adversarial robustness.
\cite{ALP} proposes to make the logit of natural samples and adversarial samples similar. 
The adversarial perturbations of input images are deemed as noises and hence \cite{xie2019feature} suggests to denoise the distorted features using non-local means or other filters to improve robustness. 
\cite{liao2018defense} proposes to deploy high-level representations to guide the denoising procedure. 
\cite{bai2021improving} observes that adversarial examples wrongly activate `negative' features which lead to the final misclassification and thus proposes Channel-wise Activation Suppressing (CAS) strategy to suppress those `negative' features to improve a model's robustness.
In parallel, \cite{cifs} has similar observations and proposes a channel-wise activation method, namely CIFS, to enhance the robustness. Besides suppressing the negative activation, they also promote the positive activation for higher accuracy. 

All these methods apply to the channel/activation level, i.e., the whole channel or activation will be suppressed or promoted by a uniform scaling. Although such uniform activation scaling (suppression or promotion) methods do improve robustness as seen from \cite{bai2021improving,cifs}, scaling uniformly, especially suppression, may lead to the information loss of the scaled activation. This inspires us to think about \textit{can we robustly scale/calibrate activation without losing their information which may help the model to further improve its robustness?}

In this paper, we propose a new and fine-grained activation scaling method to improve the robustness of CNN models, i.e., instead of scaling each activation using a uniform scaling, we conduct an Element-Wise Activation Scaling, dubbed EWAS. By means of EWAS, the distorted activation are not completely suppressed or promoted, but are re-calibrated in a fine-grained manner. 
Our key contributions are summarized as follows:

\begin{itemize}
    \item We propose the EWAS module, which can be easily added to the existing CNN models. EWAS performs activation adjustment in an element-wise fashion to improve the CNNs' robustness. The core component of EWAS is an auxiliary and class-aware classifier which is used to generate the element scaling factor. 
    \item We conduct extensive experiments to evaluate the effectiveness of EWAS in terms of adversarial robustness, where different CNN models, datasets, AT methods and adversarial attacks are deployed. The experimental results show that our EWAS-based models can greatly improve the robustness of the evaluated models over SOTA \cite{bai2021improving}\cite{cifs}. In the best case against C\&W attack, EWAS can improve the robustness by 37.65\% to 82.35\% and makes its adversarial accuracy comparable to its nature accuracy, 84.73\%.  
\end{itemize}
\textbf{Remark:} \cite{bai2021improving} and \cite{cifs} strive to minimize the activation difference between nature examples and adversarial counterparts. However, the activation analysis shows that EWAS does not follow this objective, where EWAS-modified CNNs demonstrate different activation distributions for natural and adversarial pairs. This may provide a new thought in improving the CNNs' robustness. 

\section{Related Work}
\label{sec:related}
In this section, we briefly review adversarial training methods and the adversarial defending methods relevant to EWAS.

\textbf{Adversarial Training:} 
AT \cite{PGD} is the most widely used method to improve CNNs' robustness. AT which is a data augmentation technique for adversarial defence aims to solve the following min-max optimization problem:

\begin{equation}
\label{eq:minmax}
\mathop {\min }\limits_\theta  {E_{(x,y) \sim D}}[\mathop {\max }\limits_\delta  (L(y,F(x + \delta ,\bm{\theta} )))]
\end{equation}
where $F$ represents a CNN model with weight parameters $\bm{\theta}$ and $L$ is the loss function, e.g., cross-entropy loss. $x$ and $y$ are a natural example and its corresponding label from dataset $D$. $x+\delta$ represents the adversary of $x$ with adversarial perturbation $\delta$ which is within ${l_p}$-norm distance and satisfies ${\left\| \delta \right\|_p} < \varepsilon $. Here, similar to previous methods, \cite{cifs,bai2021improving} we set $p=\infty$. The inner maximization problem aims to generate the strong adversary, while the outer minimization problem is the model training procedure to learn model weights $\bm{\theta}$ with adversarial examples.

Different adversarial attacks can be applied to AT, such as Projected Gradient Descent (PGD) \cite{PGD} and fast gradient sign method (FGSM) \cite{wong2020fast}. Since the emergence of AT, diverse methods have been proposed to improve the effectiveness and efficiency of AT.  
\cite{wong2020fast} combined FGSM \cite{fgsm} with random initialization to make FGSM applicable to AT with lower cost. \cite{mart} observed the impact of misclassified samples on models' robustness and thus proposed a misclassification-aware AT (MART) to improve the adversarial robustness. 
Although AT can improve adversarial robustness, it also sacrifices the accuracy for natural examples. Two improvements, TRADES \cite{TRADES} and FAT \cite{FAT}, are proposed to address the accuracy drop for natural examples. 

\begin{figure}[!htbp]
        \centering
            \subfigure[Channel-wise activation scaling, CAS and CIFS]
            {
                \begin{minipage}[b]{0.45\textwidth}
                \includegraphics[width=0.8\textwidth]{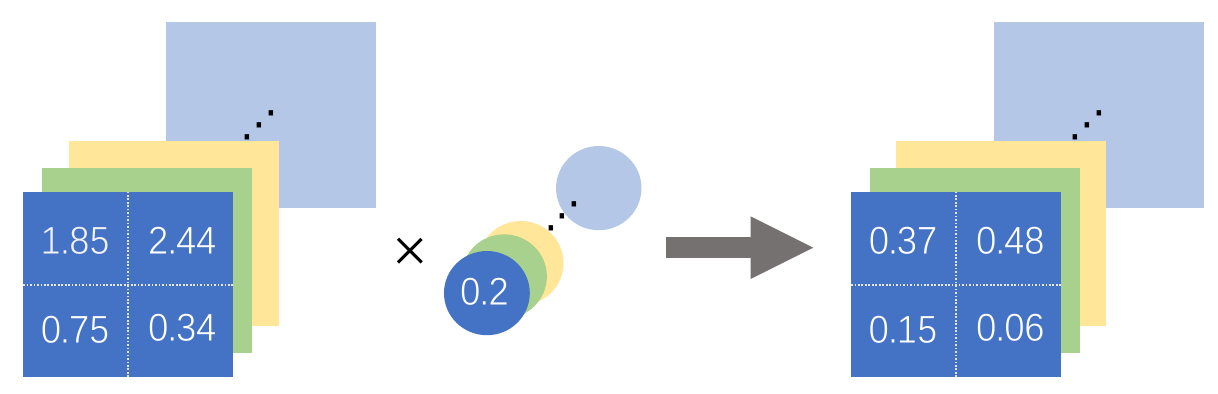}
                \label{fig:overview-cws}
                \end{minipage}
            }\\
            \subfigure[Element-wise activation scaling, EWAS ] 
            {
                \begin{minipage}[b]{0.45\textwidth}
                \includegraphics[width=0.8\textwidth]{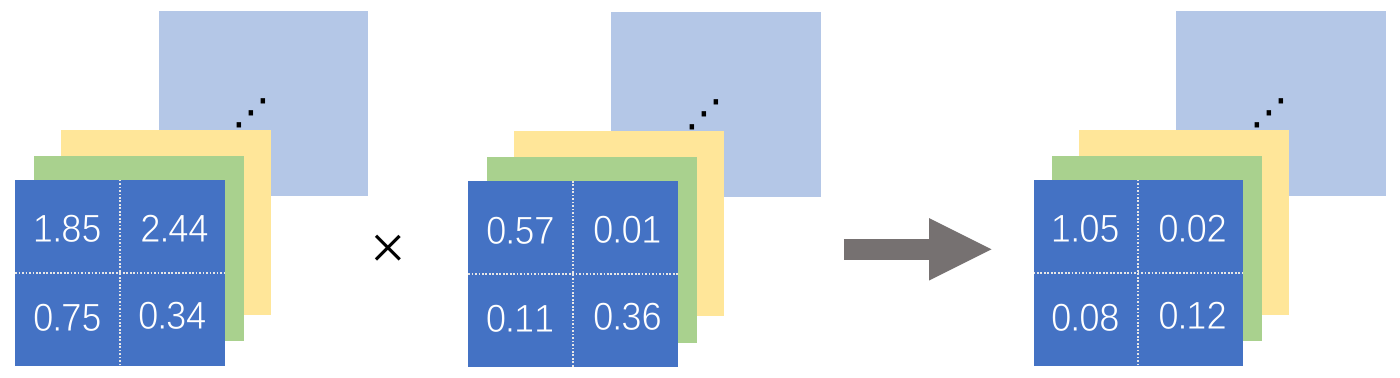}
                \label{fig:overview-ewas}
                \end{minipage}
            }
\caption{Channel-wise scaling vs element-wise scaling. Element-wise scaling conducts a more fine-grained scaling to the intermediate activation. }
\label{fig:cmp-cas}
\end{figure}

\textbf{Robust Activation Manipulation:} 
 Some works strive to understand the difference between adversarial examples and their nature counterparts from the lens of intermediate activation. 
 Then, some works propose to diminish such difference to improve the robustness, e.g., adversarial logit pairing \cite{ALP}. 
 Two concurrent works, CAS \cite{bai2021improving} and CIFS \cite{cifs}, adopt the robust activation scaling. 
\cite{bai2021improving} proposed Channel-wise Activation Suppressing (CAS) strategy to suppress redundant activation that are `\textit{negatively}' activated by adversarial examples. 
Similarly, \cite{cifs} observed that some channels, which are over-activated by adversarial examples but are not important to correct prediction, undermine the adversarial robustness.
Thus, they proposed CIFS which identifies those channels and suppresses them to improve the robustness.
 The two above-mentioned methods both feature a channel-level scaling, i.e., the whole activation is uniformly scaled as shown in Fig. \ref{fig:overview-cws}. 
 We conjecture that these uniformly scaled channels carry some useful information which can contribute to the robust prediction, so individually adjusting each element within an activation would help improve a model's robustness. 
 The idea is simple but effective as we can see from our extensive evaluation in Section \ref{sec:exp} which justifies our conjecture.    

\section{Element-Wise Activation Scaling}
Fig. \ref{fig:method} demonstrates the overview of EWAS, where EWAS is a plug-in module being added to the existing CNN models. The EWAS module is trained with the backbone network by means of an auxiliary loss function. Each layer of a CNN can be equipped with an EWAS module, but we empirically find that for a CNN model, simply adding one EWAS module demonstrates the best adversarial robustness. 
Next, we proceed to the module and how to train it.

\subsection{EWAS Module}

Let $z^l \in \mathbb{R}^{C\times H \times W}$ denote the activation of layer  $l$ which has an EWAS module, where $C$ denotes the number of channels and $H$ and $W$ are the height and width, respectively. Each element in $z^l$ is expected to have an individual scaling factor and thus we have $\bm{m}\in \mathbb{R}^{C\times H \times W}$ to denote the scaling factor vector. As seen in \cite{bai2021improving}\cite{cifs}, class-
related activation modification is instrumental in improving the robustness. Hence, we also deploy an auxiliary classifier to have the class-related feature and determine the element-wise scaling factor $\bm{m}$.

\subsubsection{Auxiliary Linear classifier (ALC)}
\begin{figure}[!htbp]
        \centering
        \includegraphics[width=0.45\textwidth]{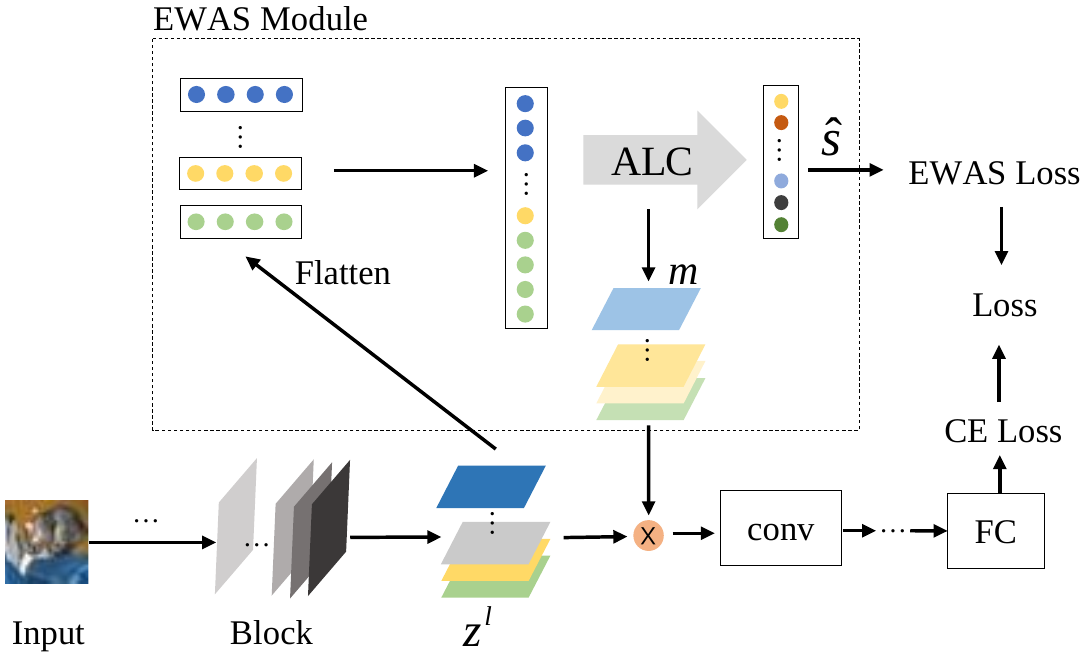}  
\caption{Three steps of EWAS: 1) Flatten $z^l$  and input it into ALC, and the output score of ALC $\hat s$ calculate EWAS loss. 2) Class Related Scaling (CRS): $\bm{m}$ from ALC's weight element-wise multiply with $z^l$ to scaling the $z^l$ to get $\tilde z^l$. 3) Forward $\tilde z^l$ into the model's next layer. }
\label{fig:method}
\end{figure}

The core of EWAS is the scaling factor $\bm{m}$. A good scaling factor $\bm{m}$ will suppress redundant and negative elements while retaining or promoting robust and positive elements. Inspired by CAS \cite{bai2021improving}, we add an auxiliary linear classifier (ALC) to the original model and use ALC to derive $\bm{m}$. 
The overview of EWAS can be seen in Fig. \ref{fig:method}. ALC takes activation $z^l$ as the input and outputs classification scores of $K$ classes. 

Let $\theta^\text{ALC} \in \mathbb{R}^{C \cdot H \cdot W\times K}$ denote the parameters of ALC. ALC parameters $\theta^\text{ALC}$ are deployed to generate the scaling mask $\bm{m}$. ALC is a class-related scaling classifier, i.e., we have $\theta^\text{ALC}_k \in \mathbb{R}^{C\cdot H \cdot W}$ to represent the parameters related to class $k$ and $\theta^\text{ALC}_k$ will be converted to the scaling factor $\bm{m}$. 
In the training stage, the ground truth label $y$ serves as the class index to select which class' parameters to update. In the inference stage, since there is no label information provided, the maximum value of $\hat s$ predicted by ALC is used as the class index. The scaling factor $\bm{m}$ is formulated as follows:

\begin{equation}
\label{eq:CRS}
\bm{m} = 
\begin{cases}
\text{reformat}(\theta _y^{{\rm{ALC}}}),\quad & {\text{ (training  stage)}} \\
\text{reformat}(\theta _{\arg \max (\hat s)}^{{\rm{ALC}}}),\quad &{\text{ (inference stage)}}
\end{cases}
\end{equation}
Note that the scaling factor $\bm{m}$ is reformatted into size $\mathbb{R}^{C\times H \times W}$. After obtaining the scaling factor $\bm{m}$, we perform element-wise multiplication on ${z^l}$ to obtain the adjusted activation $\hat{z}^l$. 
\begin{equation}
\label{eq:multiply}
{\tilde z^l} = {z^l} \otimes \bm{m}
\end{equation}
where $\otimes$ represents the element-wise multiplication. 
The modified activation $\hat{z}^l$ is forward-propagated to the next layer.

\subsection{Model Training}

EWAS module should be adversarially trained with the backbone network. We can add multiple EWAS modules to a CNN model, but we empirically find that adding one module shows the best robustness. We conjecture the rational behind is that fine-grained modification effectively identifies the error or negative elements. As soon as the negative elements are adjusted accordingly, more EWAS modules are not 
helpful. However, the position of EWAS is critical for the robustness and we evaluate this in Section \ref{exp:ablate}.  
Following the min-max optimization in Eq. (\ref{eq:minmax}), the EWAS-modified optimization problem can be written as:

\begin{equation}
\small
\label{eq:opt-all}
\mathop {\min }\limits_\theta  {E_{(x,y) \sim D}}[\mathop {\max }\limits_\delta  (L(y,F(x + \delta ,\theta )) + \lambda  \cdot L_\text{EWAS}(y,\hat s))]
\end{equation}
where $\hat s=\text{ALC}({f^l}(x + \delta ),\theta^\text{ALC} )$, and $f^l$ indicates the output of layer $l$. $\lambda$ is a trade-off coefficient to balance the contribution of ALC loss. $L_\text{EWAS}$ here is the same loss function as the maximization problem in Eq. (\ref{eq:minmax}), which for AT is:

\begin{equation}
\label{eq:ewas}
   L_\text{EWAS}= L_\text{CE}^\text{ALC}(\hat p(x+\delta),y)
\end{equation}



EWAS can combine with diverse adversarial training methods such as TRADES \cite{TRADES}, MART (\cite{mart}), and the EWAS loss function needs to be modified accordingly.  More details of different loss functions can be seen in Appendix \ref{apdx:lewas}. The training algorithm is given in Appendix \ref{apdx:algo}.

\section{Experiments}
\label{sec:exp}

In this section, we extensively evaluate the effectiveness of EWAS in terms of adversarial robustness in comparison with the state of the arts \cite{bai2021improving}\cite{cifs}.
We use WideResNet-32-10 (we call it WideResNet), WideResNet-28-10 and ResNet-18 as in CAS and CIFS, and train models using CIFAR10 \cite{cifar10} and SVHN \cite{svhn} datasets. We empirically determine the best layer to add the EWAS module, i.e., the 15th layer for ResNet-18 , the 19th layer of WideResNet-28-10 and 25th layer for WideResNet, respectively. 
AT \cite{PGD} and its variants MART \cite{mart} and TRADES \cite{TRADES} are used to train the models with EWAS-modified models, and three white-box attack methods are considered, FGSM \cite{fgsm}, PGD-20 \cite{PGD}, C\&W \cite{c&w}. All attacks are perturbed by ${l_\infty }$-norm with bound $\epsilon= 8/255$ and step size $\epsilon/4$. Note that to train the model with MART and TRADES, we need to modify the loss function accordingly. Models are trained for 120 epochs under AT, and the setting for other AT variants can be seen in Appendix \ref{apdx:exp}.
We also visualize how the EWAS module affects the intermediate activation in Appendix \ref{apdx:pewas} and investigate how $\lambda$ effect attack evaluation results in Appendix \ref{apdx:evlam}.

\subsection{Robustness Analysis and Evaluation}

In this section, we first evaluate EWAS against CAS and CIFS, which are closest to our work.

\subsubsection{Robustness Evaluation}
$\lambda$ in Eq. (\ref{eq:opt-all}) is a critical parameter for EWAS module training, and the two datasets have different values, 0.01 for CIFAR10 and 0.05 for SVHN. Later, in the ablation study, we further evaluate the impact of $\lambda$.
The adversarial accuracy of the last epoch is reported for each model.


\begin{table}[h!]
\small
\begin{tabular}{ccccc}
\Xhline{3\arrayrulewidth}
\begin{tabular}[c]{@{}c@{}}ResNet-18\end{tabular} & Natural & FGSM  & PGD-20 & C\&W  \\ \Xhline{3\arrayrulewidth}
AT                                                       & 84.47   & 61.09 & 44.33  & 44.70 \\
AT+CAS                                                   & \textbf{85.89}   & 61.17 & 50.55  & 52.56 \\
AT+CIFS                                                  & 82.70   & 58.10 & 49.49  & 50.24 \\
AT+EWAS                                                   & 84.73   & \underline{\textbf{ 65.78 }} & \underline{\textbf{ 64.84 }}  & \underline{ \textbf{82.35} } \\ \hline
TRADES                                                   & 79.57   & 62.26 & 52.29  & 49.18 \\
TRADES+CAS                                               & \textbf{83.05}   & \textbf{63.81}  & 56.63  & 60.03 \\
TRADES+EWAS                                               & 80.35     & 61.85 & \textbf{61.29}   &\textbf{74.92 } \\ \hline
MART                                                     & 78.86   & 61.87 & 51.61  & 46.97 \\
MART+CAS                                                 & \underline{ \textbf{86.40} }   & 62.61 & 54.33  & 61.49 \\
MART+EWAS                                                 & 81.80  & \textbf{65.31} & \textbf{64.01}  & \textbf{79.67} \\ \Xhline{3\arrayrulewidth}\\
\Xhline{3\arrayrulewidth}
WideResNet-28-10 & Natural              & FGSM                 & PGD-20               & C\&W \\ \Xhline{3\arrayrulewidth}
AT &      87.29         &      58.50            &    49.17            &    48.68             \\ 
AT+CAS &    \textbf{88.05}        &      57.94            &    49.03            &    49.97        \\ 
AT+CIFS &     85.56      &     61.34          &   53.74            &    53.20       \\
AT+EWAS &     85.29     &     \textbf{62.23} &   \textbf{55.66}           &    \textbf{67.07}       \\
\Xhline{3\arrayrulewidth}\\
\Xhline{3\arrayrulewidth} 
WideResNet & Natural              & FGSM                 & PGD-20               & C\&W \\ \Xhline{3\arrayrulewidth}
AT         & 86.65 & 63.71                & 47.06                & 45.75                \\
AT+EWAS     & \underline{ \textbf{87.12} }             & \textbf{64.05}       & \textbf{59.90}       & \textbf{73.01}       \\
\hline
TRADES     & \textbf{84.16}       & \underline{ \textbf{65.34} } & 52.92                & 51.61                \\
TRADES+EWAS &  83.96              &   64.50        & \textbf{62.39}  &  \textbf{74.88}  \\ \hline
MART       & \textbf{84.39}       & \textbf{65.10}       & 50.39                & 48.77                \\
MART+EWAS   &   80.84      &     63.19            & \underline{ \textbf{65.40} }       & \underline{ \textbf{76.72} }  \\
\Xhline{3\arrayrulewidth} 
\end{tabular}
\caption{Robustness (accuracy (\%) on various white-box attacks) comparison of defense methods on CIFAR10. The best results are marked with an underline.}
\label{tbl:c10-rc}
\end{table}

\begin{table}[!htbp]
\centering
\begin{tabular}{ccccc}
\Xhline{3\arrayrulewidth}
ResNet-18      & Natural        & FGSM           & PGD-20         & C\&W           \\ \Xhline{3\arrayrulewidth}
AT        & 93.72 & 65.87 & 50.35          &   47.89        \\
AT+CAS & \textbf{94.08} & 65.24 & 48.47 &46.15 \\
AT+CIFS & 93.94 & 66.24 & 52.02 &50.13\\
AT+EWAS  & 92.18 & \textbf{71.57} & \textbf{59.01}  &\textbf{69.67} \\
\Xhline{3\arrayrulewidth}
\end{tabular}
\caption{Experimental results for SVHN.}
\label{tbl:svhn-rc}
\end{table}

Table \ref{tbl:c10-rc} shows the experimental results for CIFAR10. As see from Table \ref{tbl:c10-rc}, 
EWAS greatly improves the robustness of models, especially the robustness against PGD and C\&W attacks. The robust accuracy of ResNet-18 against C\&W increases by 37.65\% under AT, and such huge improvement makes its robust accuracy comparable to its natural accuracy, where the difference is only 2.38\%. 
Also for PGD attack, EWAS significantly improves the adversarial accuracy by up to 20.51\%. 
Although MART and TRADES can improve the robustness, the vanilla AT achieves the best robustness for ResNet-18 under CIFAR10. 
For WideResNet-28-10, EWAS outperforms CAS and CIFS in terms of robust accuracy under three attacks, but CAS achieves the best natural accuracy.
For WideResNet, MART and TRADES demonstrate better performance than the vanilla AT, where we obtain the best robust accuracy under MART. 
Table \ref{tbl:svhn-rc} summarizes the results for SVHN, where EWAS performs superiority over CAS and CIFS in terms of the adversarial accuracy, and the improvement against C\&W is up to 19.54\%. 

\begin{table}[!htbp]
\centering
\begin{tabular}{cccc}
\Xhline{3\arrayrulewidth}
ResNet-18       & Vanilla & CAS   & EWAS   \\ 
Robust Accuracy & 39.35   & 65.31 & 63.22
\\ \Xhline{3\arrayrulewidth}
\end{tabular}
\caption{Robustness accuracy against AutoAttack on CIFAR10. CIFS does not report this.}
\label{tbl:aa}
\end{table}

We also evaluate the robust accuracy against AutoAttack \cite{croce2020reliable} as \cite{bai2021improving}, which is a parameter-free attacks framework consist of both white-box and black-box attack. We use the AutoAttack including one white-box attack (APGD-DLR \cite{croce2020reliable}) and one black-box attack (Square Attack \cite{squareattack}). 
As shown in Table \ref{tbl:aa}, EWAS can improve the robustness of DNN but 2.19\% lower than CAS.


\subsubsection{Feature Analysis}

\begin{figure*}[h!]
        \centering
            \subfigure[AT] 
            {\label{subfig:at}
                \begin{minipage}[b]{0.2\textwidth}
                \includegraphics[width=1\textwidth]{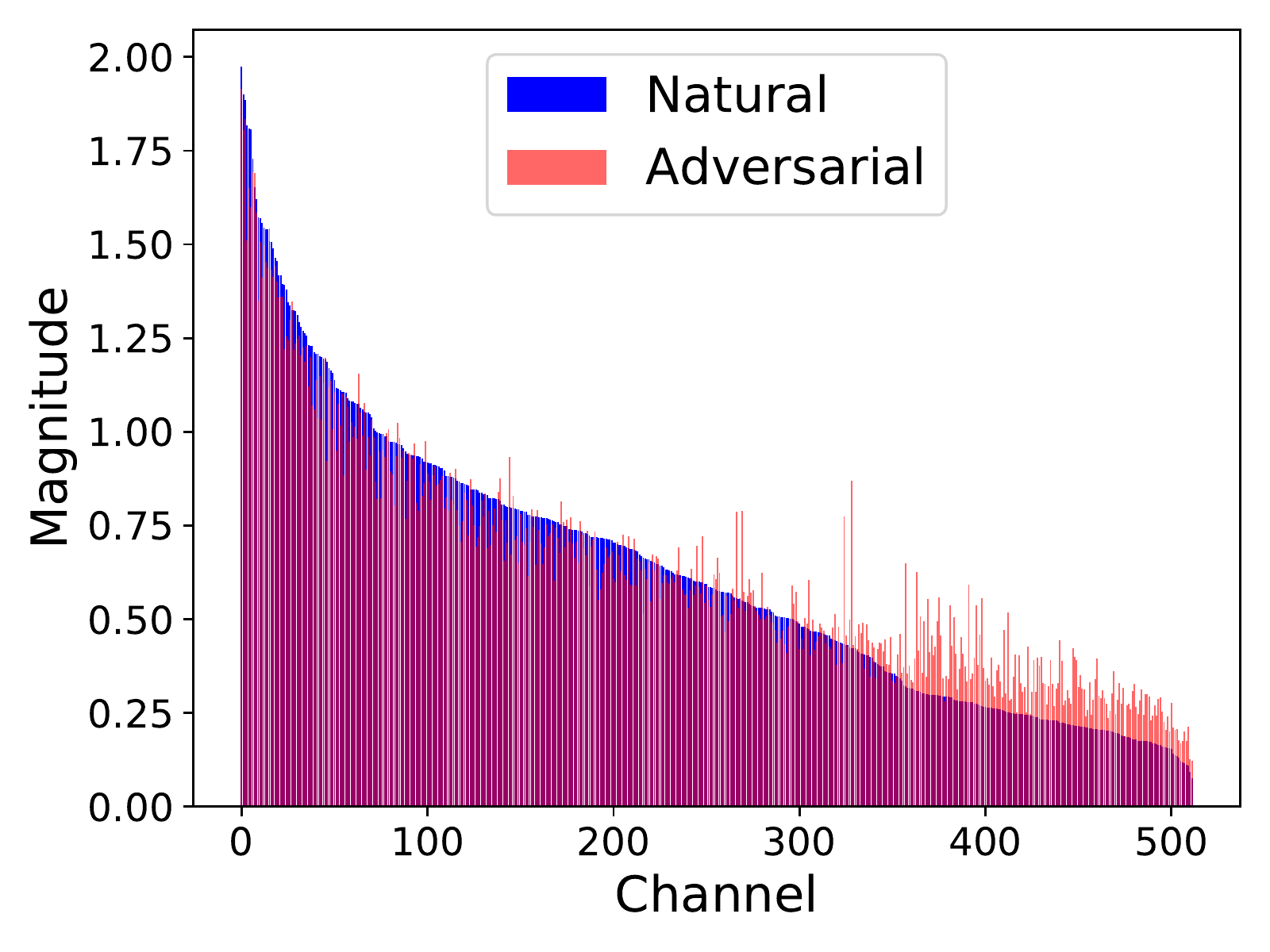}
                \includegraphics[width=1\textwidth]{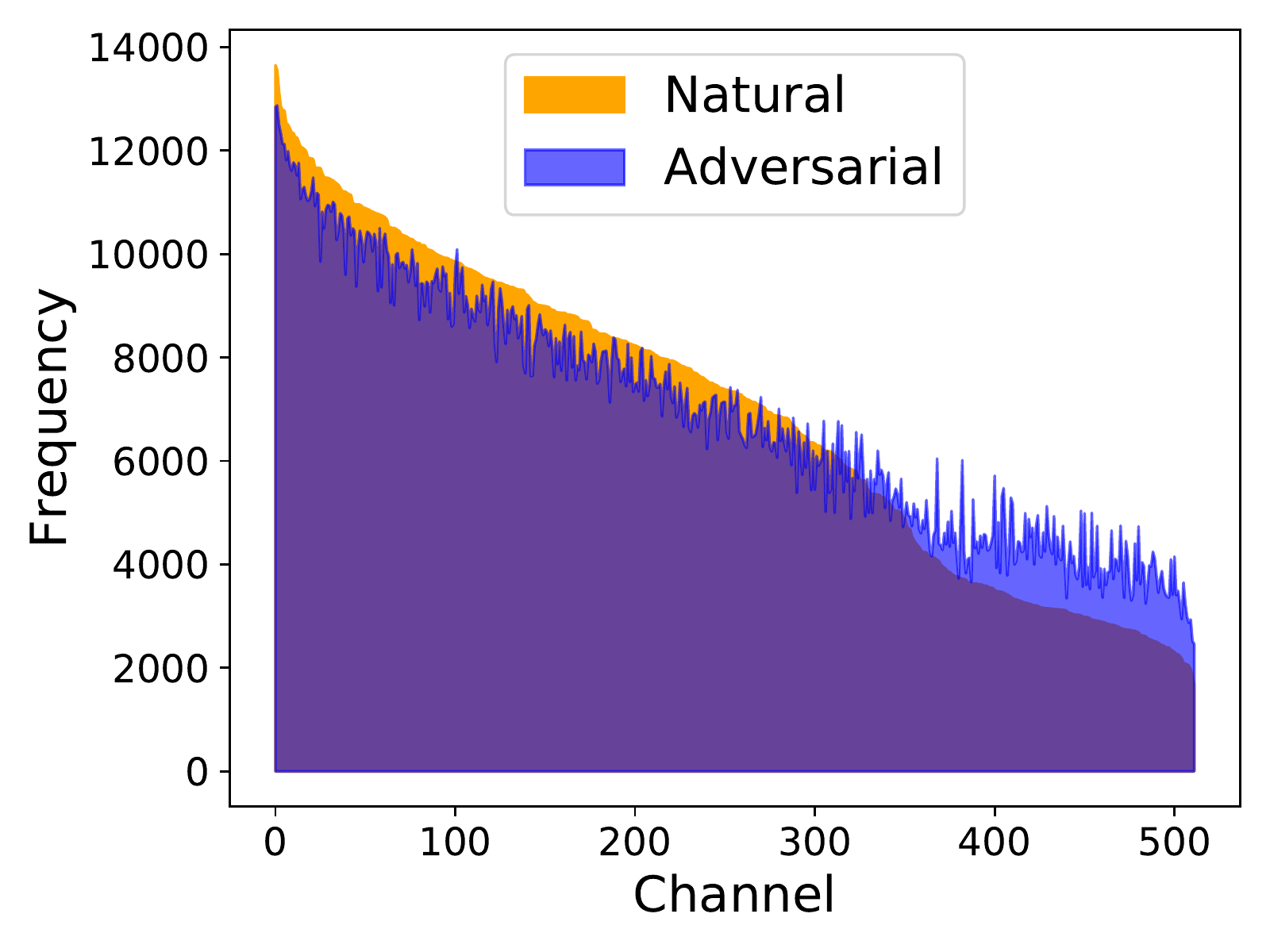}
                \end{minipage}
            }
           \subfigure[CAS] 
            {\label{subfig:cas}
                \begin{minipage}[b]{0.2\textwidth}
                \includegraphics[width=1\textwidth]{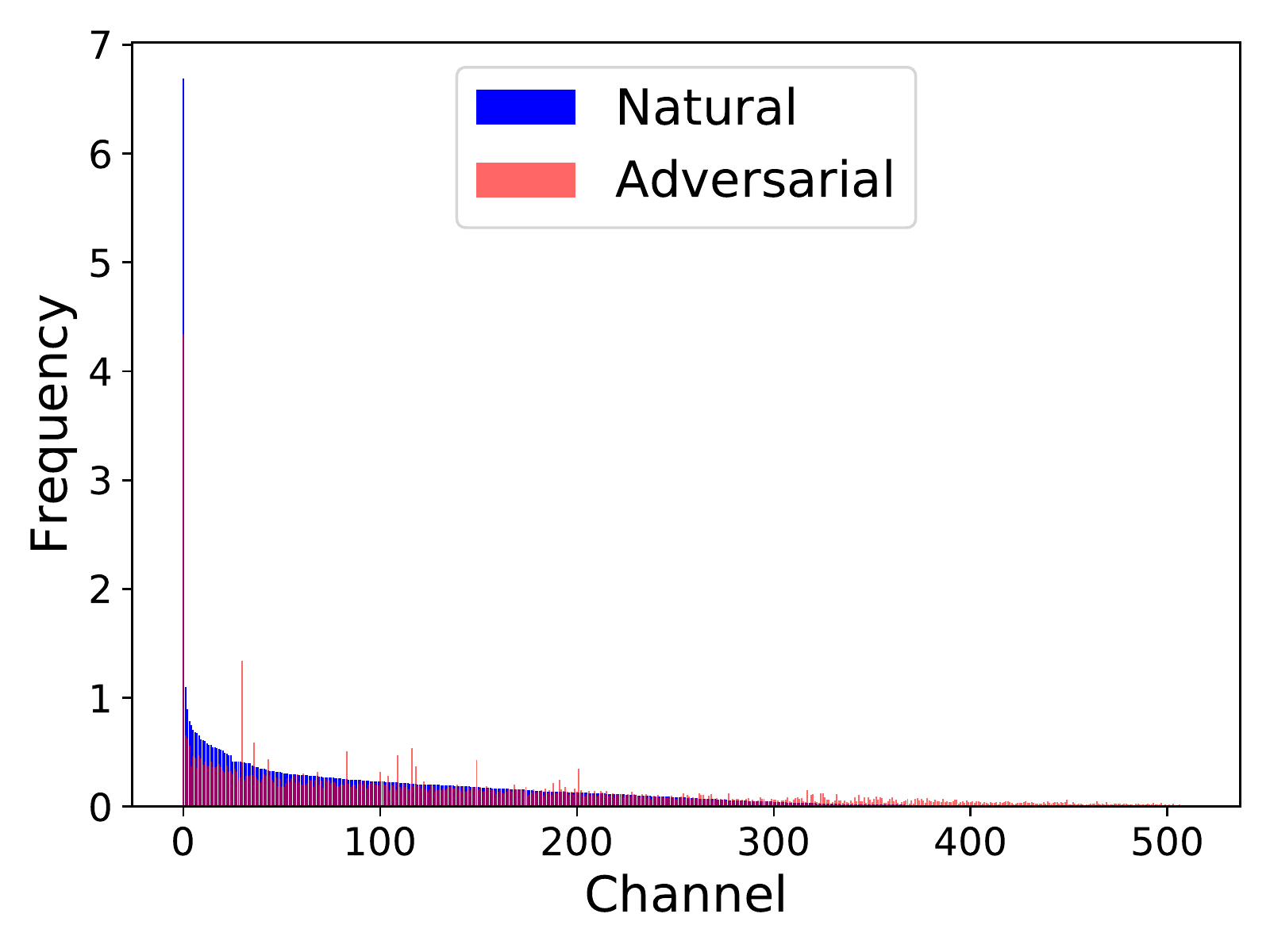}
                \includegraphics[width=1\textwidth]{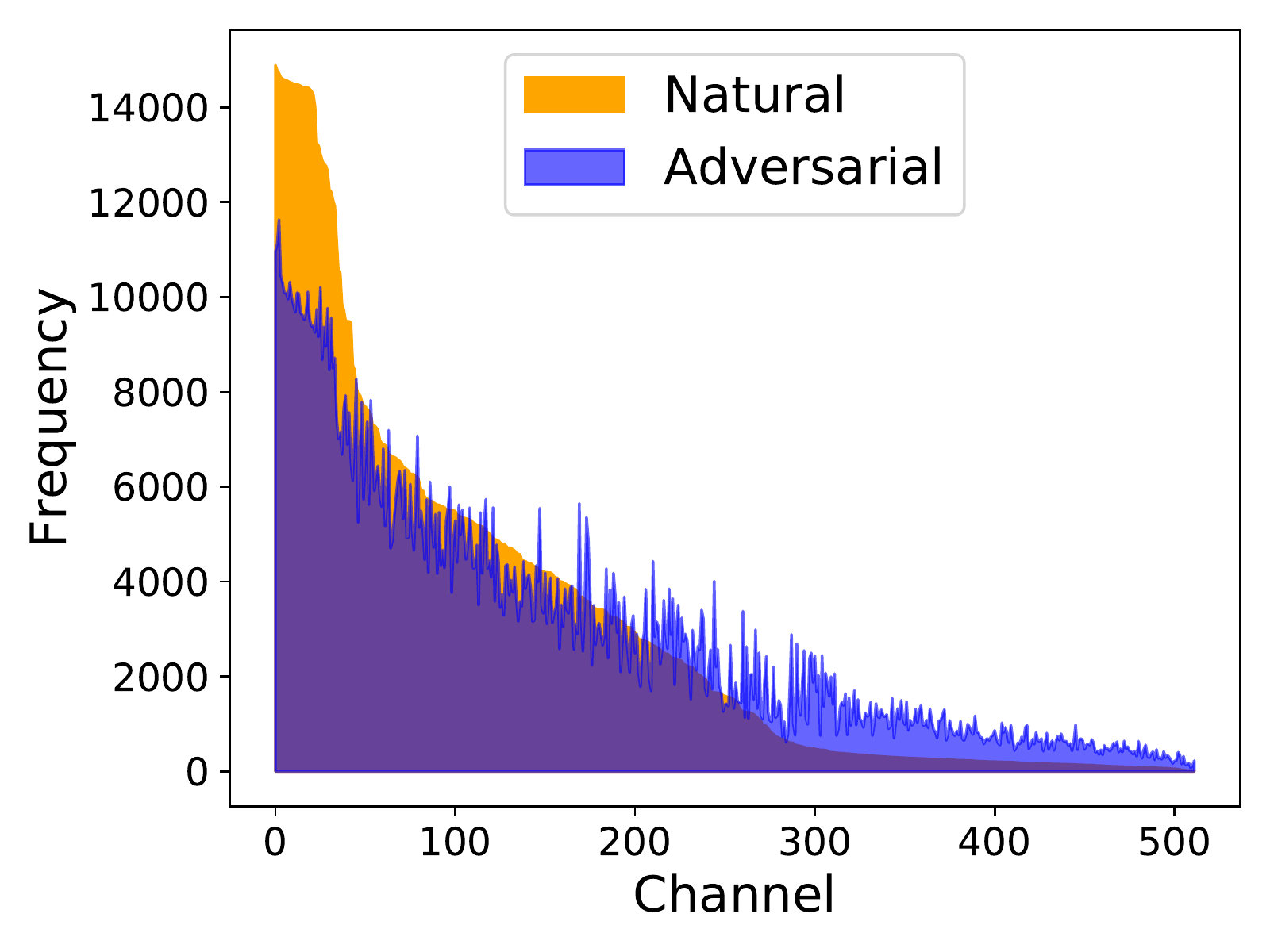}
                \end{minipage}
            }
            \subfigure[CIFS] 
            {\label{subfig:cifs}
                \begin{minipage}[b]{0.2\textwidth}
                \includegraphics[width=1\textwidth]{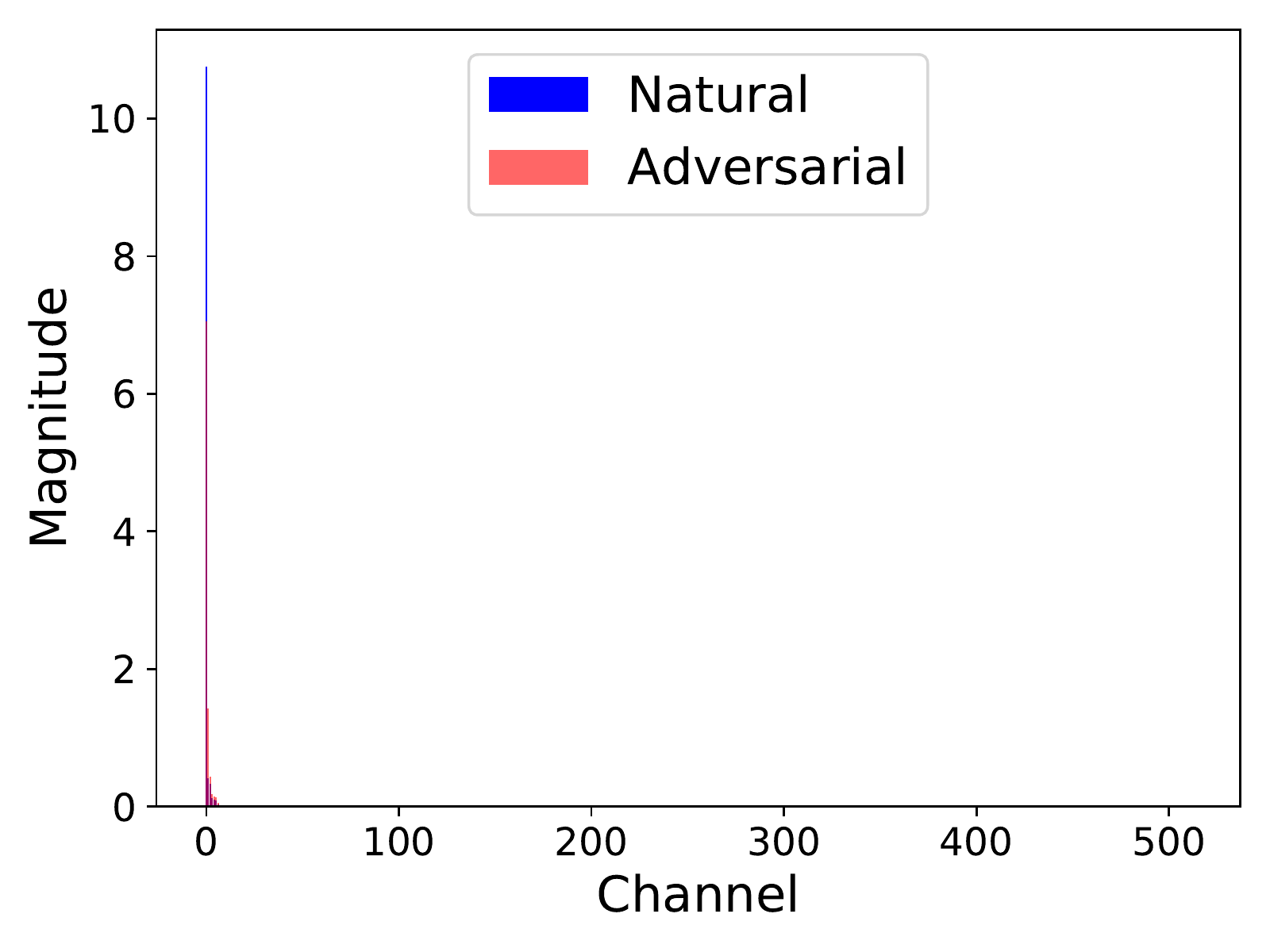}
                \includegraphics[width=1\textwidth]{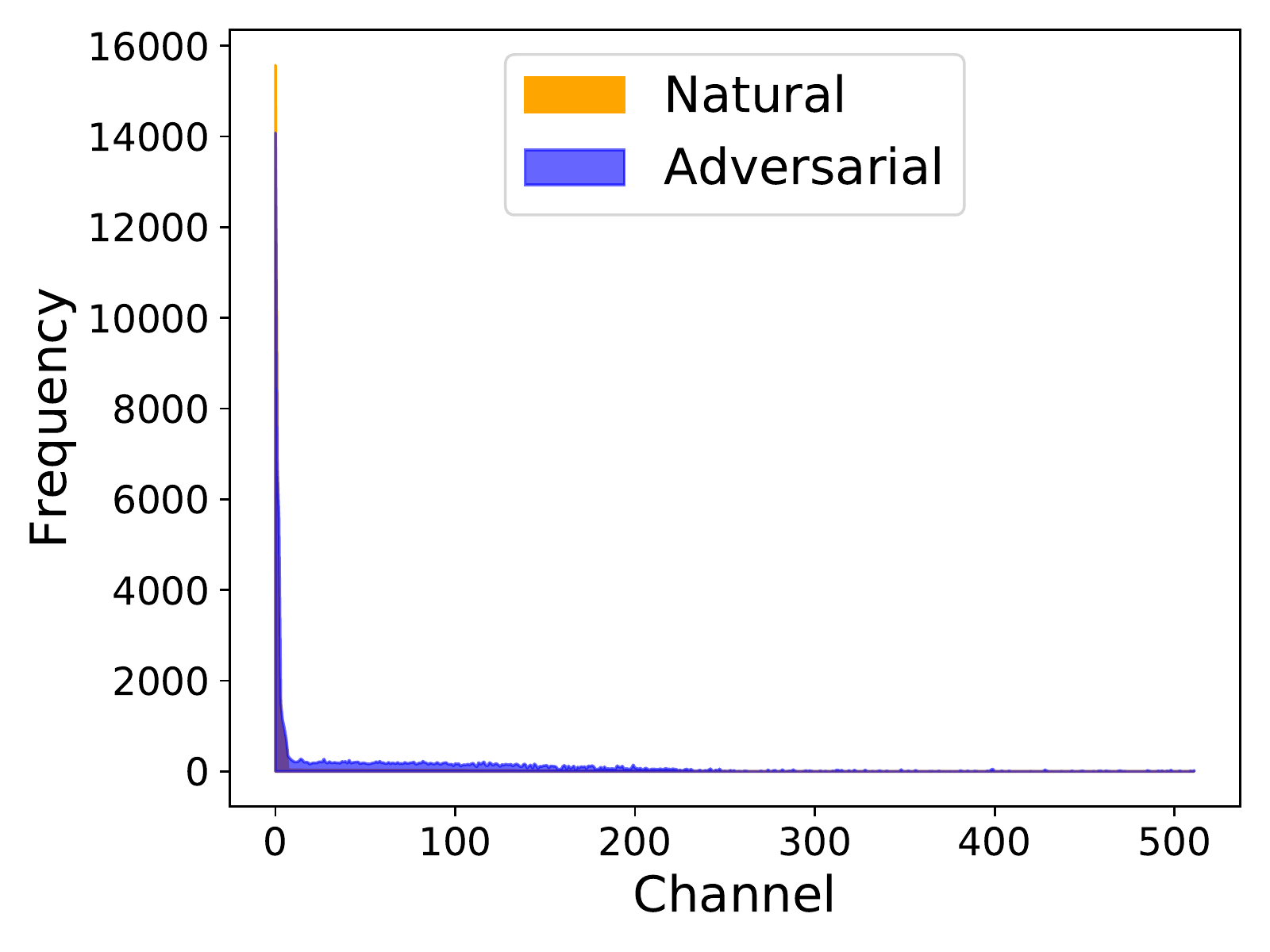}
                \end{minipage}
            }
            \subfigure[EWAS(our)] 
            {\label{subfig:ewas}
                \begin{minipage}[b]{0.2\textwidth}
                \includegraphics[width=1\textwidth]{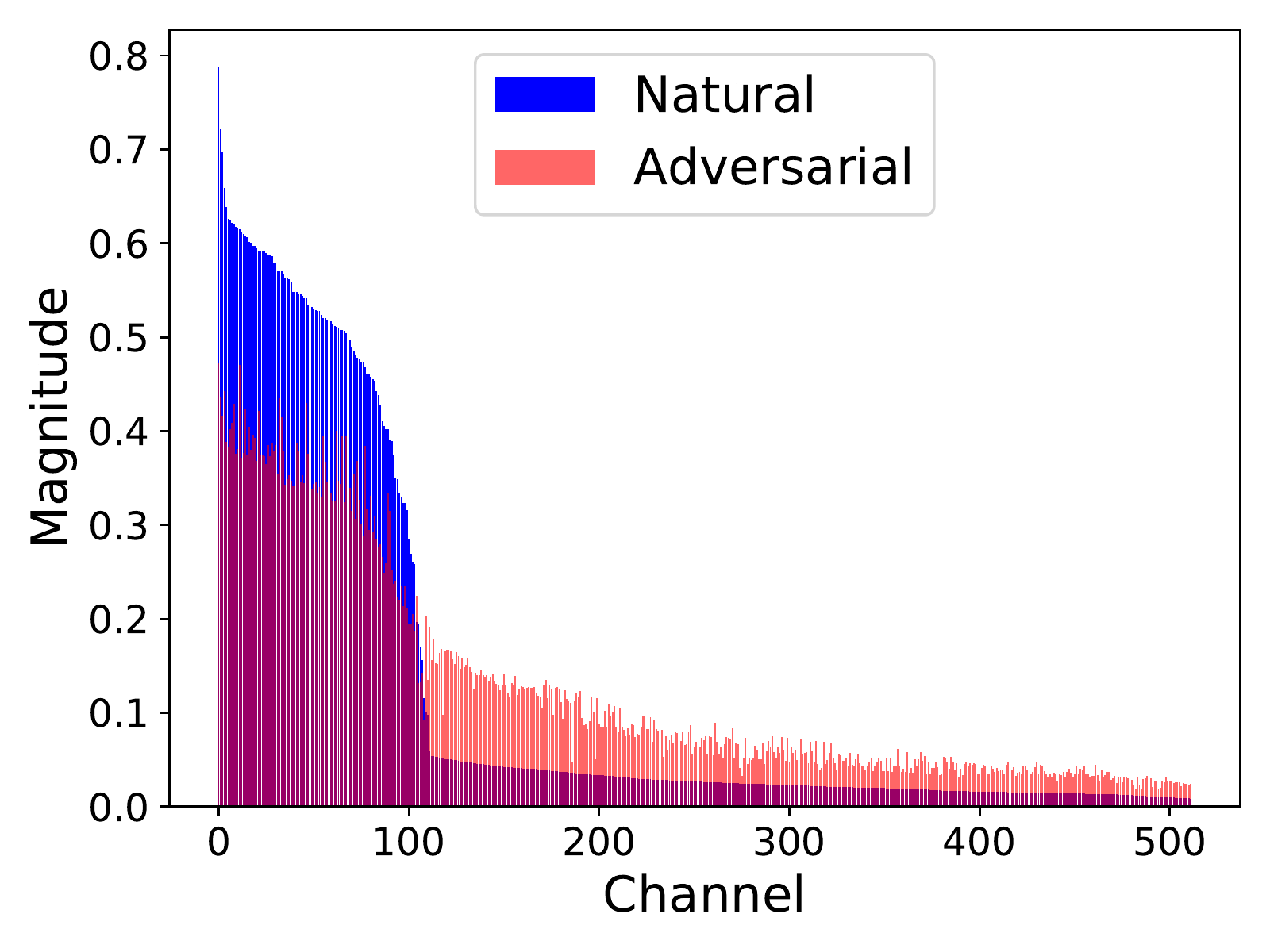}
                \includegraphics[width=1\textwidth]{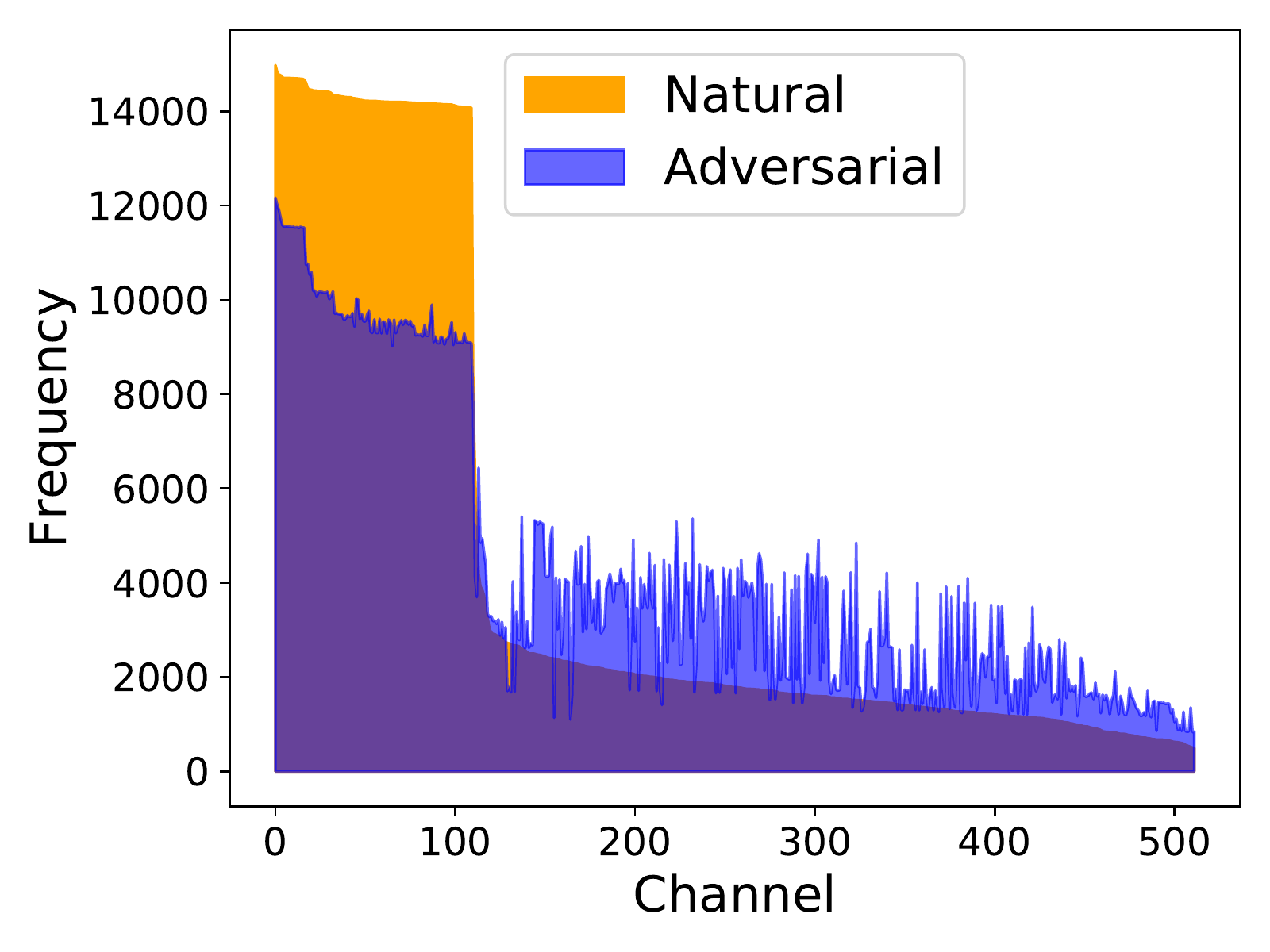}
                \end{minipage}
            }

\caption{Comparison of activation magnitude and frequency between adversarial and natural samples on different defense methods. Natural samples are from CIFAR-10 "airplane" class.}
\label{fig:compare}
\end{figure*}

We visualize the activation of the penultimate layer (the last convolutional layer) of ResNet-18 w.r.t the activation magnitude and frequency in Fig. \ref{fig:compare}, and the visualization details are shown in Appendix \ref{apdx:DVAF}. As observed from the figure, the 4 methods demonstrate significantly different results. AT, CAS and CIFS aim to make adversarial examples similar to natural examples, whereas EWAS presents more difference between natural examples and adversarial examples. EWAS tends to have different activation distributions for two types of examples. This may provide a new direction to improve CNNs' robustness. The visualization of WideResNet activation is shown in Appendix \ref{apdx:vw}.


\subsection{Ablation Study}
\label{exp:ablate}

\subsubsection{The Impact of $\lambda$}

In this part, we evaluate the impact of $\lambda$ in Eq. (\ref{eq:opt-all}). 
We train EWAS-modified ResNet-18 with 6 different values $\lambda=[0.01,0.05,0.1,0.5,1,2]$ under AT on CIFAR10 and SVHN. 
$\lambda$ serves two roles in the model training: 1) it balances the contributions of the backbone classifier and the auxiliary classifier; 2) it controls the strength of element scaling. The results are reported in Table \ref{tbl:dflc} and Table \ref{tbl:dfls}.



For CIFAR10, the natural and robust accuracies decrease with the increase of $\lambda$ over different attacks. 
When $\lambda$ (i.e. $\lambda=2$) is large, the model training cannot be converged, thereby leading to low accuracy for both natural and adversarial accuracy. However, for SVHN, there is no winning $\lambda$ for diverse attacks. For PGD and C\&W, the best $\lambda$ is 0.05, where $\lambda=2$ is the best for FGSM.  The best $\lambda$ for CIFAR10 is the worst selection for SVHN. 
Therefore, we choose $\lambda=0.01$ for CIFAR10, and $\lambda=0.05$ for SVHN.

\begin{table}[h!]
\centering
\begin{tabular}{ccccc}
\Xhline{3\arrayrulewidth}
$\lambda$    & Natural                     & FGSM                         & PGD-20                       & C\&W                         \\ \Xhline{3\arrayrulewidth}
0.01 & 84.73                      & 65.78                       & 64.84                       & 82.35                       \\
0.05 & 84.79                      & 63.54                       & 58.58                       & 72.64                       \\
0.1  & 84.67                      & 62.09                       & 53.77                       & 60.83                       \\
0.5  & 83.96 &  61.34 &  48.73 & 52.59 \\
1    & 83.61                      & 61.77                       & 47.45                       & 49.3                        \\
2   & 10.00 &  10.00 & 10.00 & 10.00                                                                                        \\ \Xhline{3\arrayrulewidth}
\end{tabular}
\caption{Robust comparison of different $\lambda$ on CIFAR10 for ResNet-18. The accuracies(\%) for natural and adversarial data are reported.}
\label{tbl:dflc}
\end{table}

\begin{table}[h!]
\centering
\begin{tabular}{ccccc}
\Xhline{3\arrayrulewidth}
$\lambda$ & Natural  & FGSM     & PGD-20   & C\&W     \\ \Xhline{3\arrayrulewidth}
0.01   & 19.58 & 19.58 & 19.58  & 19.58           \\
0.05   & 92.18 & 71.57 & 59.01  & 69.67 \\
0.1    & 92.72 & 72.42 & 58.38 & 63.36 \\
0.5    & 93.20 & 74.03 & 57.07 & 55.37 \\
1      & 93.02 & 74.23 & 57.30 & 54.57 \\
2      & 93.34 & 75.42 & 58.85 & 55.23 \\ \Xhline{3\arrayrulewidth}
\end{tabular}
\caption{Robust comparison of different $\lambda$ on SVHN for ResNet-18.}
\label{tbl:dfls}
\end{table}


\subsubsection{The Impact of EWAS position}

In this part, we evaluate the effect of EWAS' position on models' robustness, where we insert the EWAS module to different layers. The natural and robust accuracies against PGD-20 at different positions are shown in Fig. \ref{fig:dflayer}, and more different layers robust evaluation shown in Appendix \ref{apdx:dflayer}. The experimental results show that the best position is the first conv layer of the last block within a model. 

We think there are two reasons behind. Since the adversarial perturbation is gradually amplified along its forward propagation \cite{liao2018defense}, adding EWAS module to early layers cannot effectively discern perturbations. In addition, features in early layers are more class-agnostic, so the auxiliary classifier 
may not take effect in this case. Therefore, we empirically choose to insert the EWAS module after the 15th layer of ResNet-18, the 19th layer of WideResNet-28-10 and the 25th layer of WideResNet.
\begin{figure}[h!]
        \centering
            \subfigure[ResNet-18] 
            {\label{subfig:resnet}
                \begin{minipage}[b]{0.22\textwidth}
                \includegraphics[width=1\textwidth]{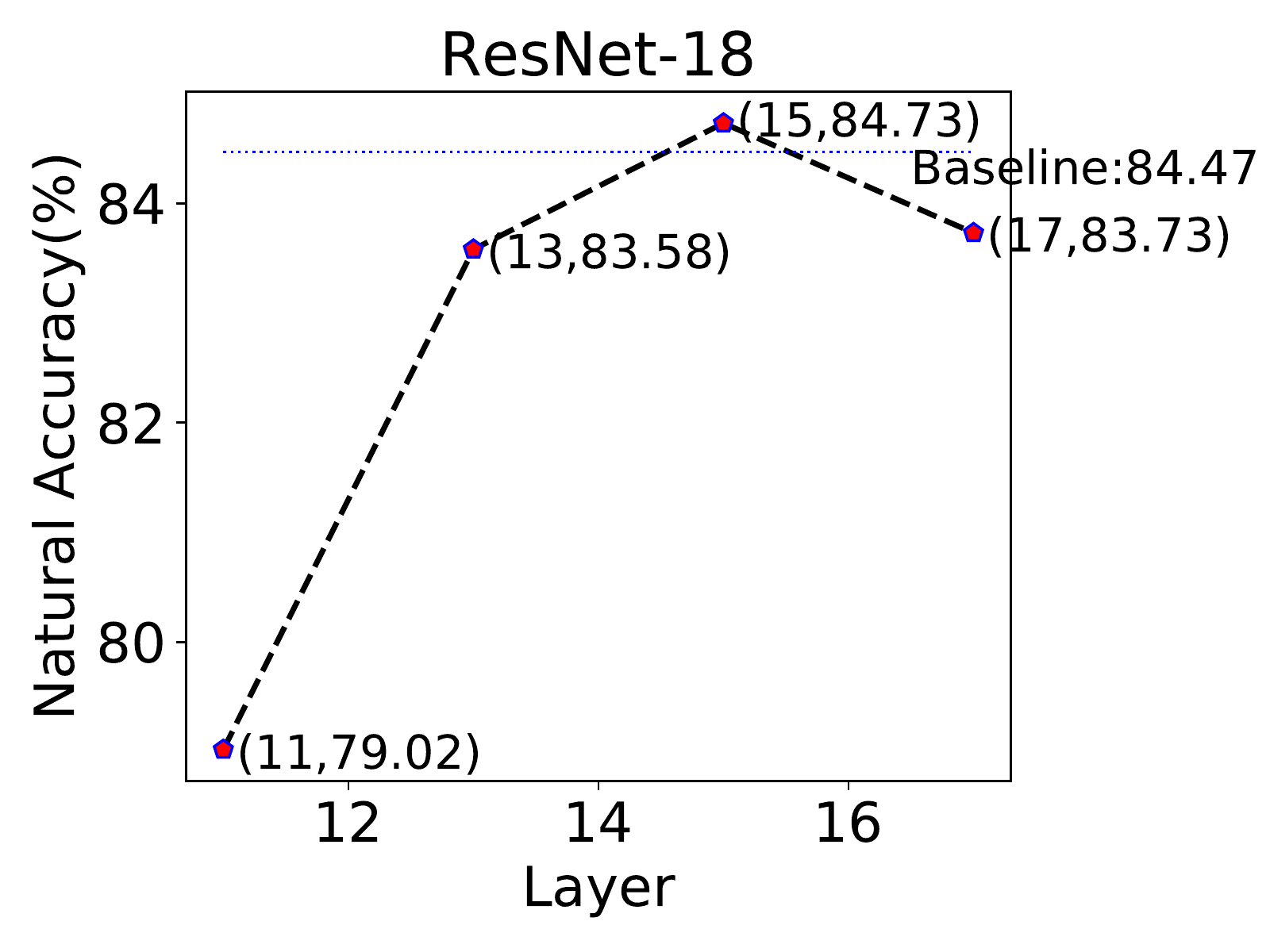}
                \includegraphics[width=1\textwidth]{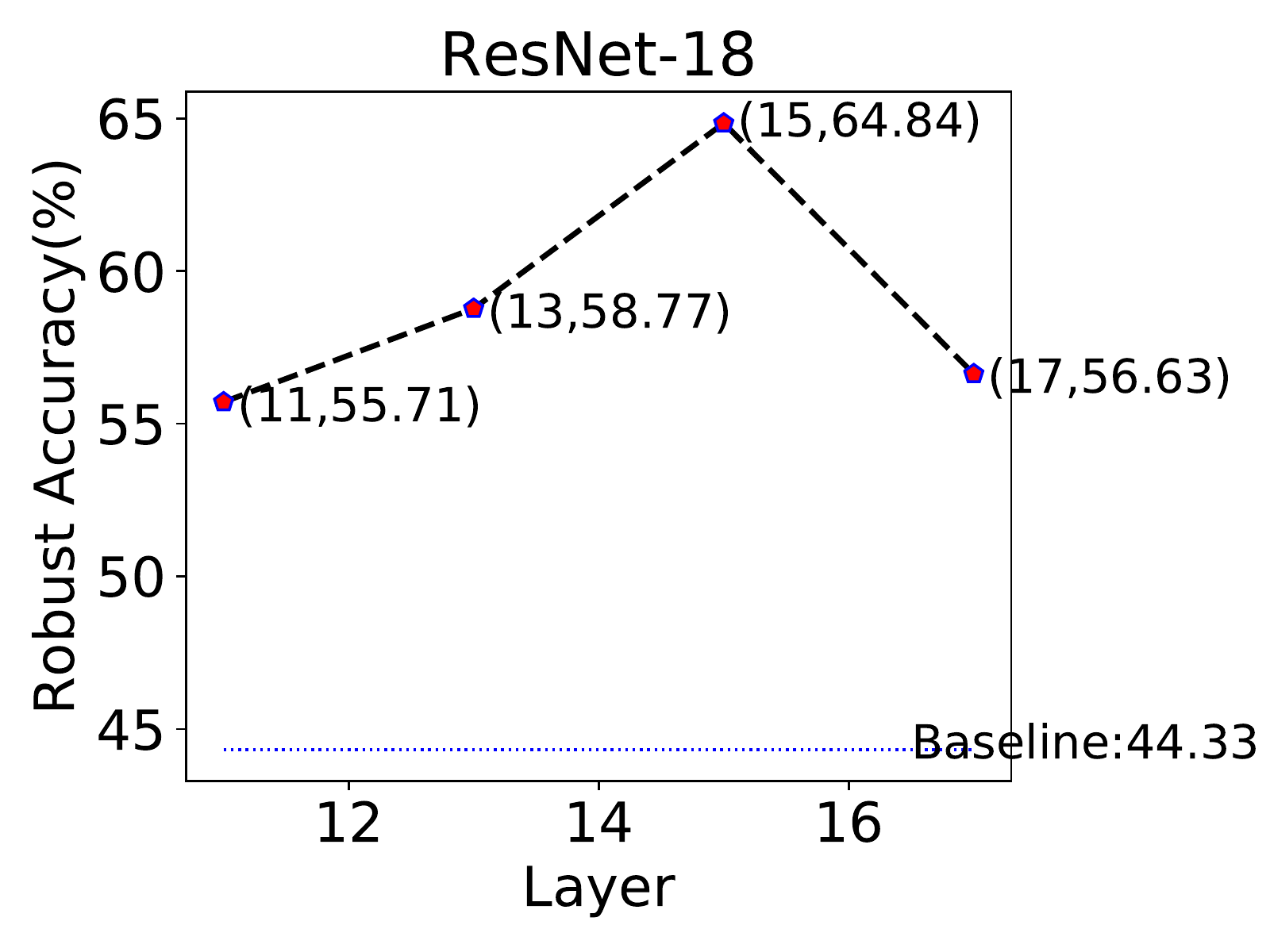}
                \end{minipage}
            }
           \subfigure[WideResNet] 
            {\label{subfig:wrn}
                \begin{minipage}[b]{0.22\textwidth}
                \includegraphics[width=1\textwidth]{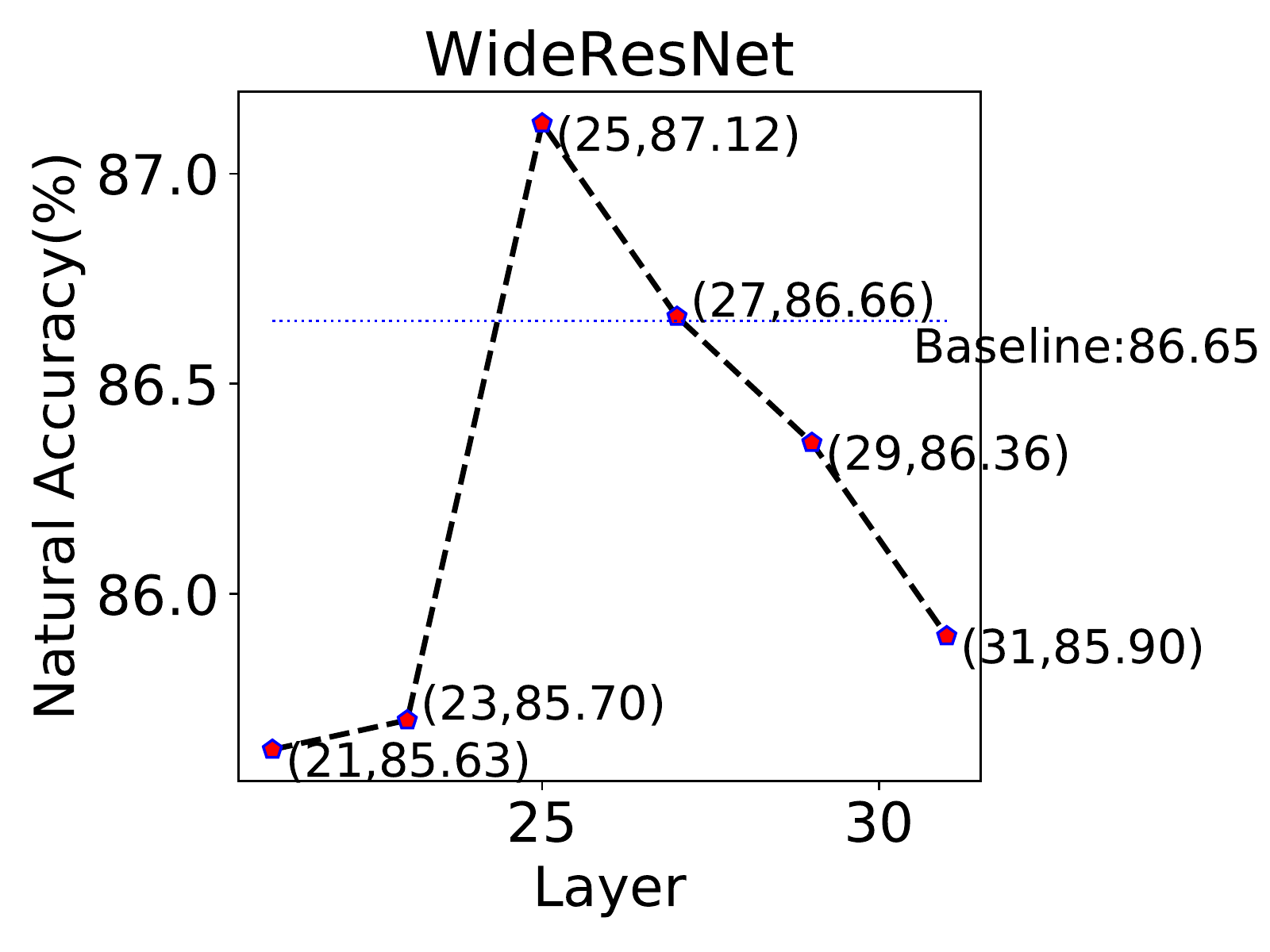}
                \includegraphics[width=1\textwidth]{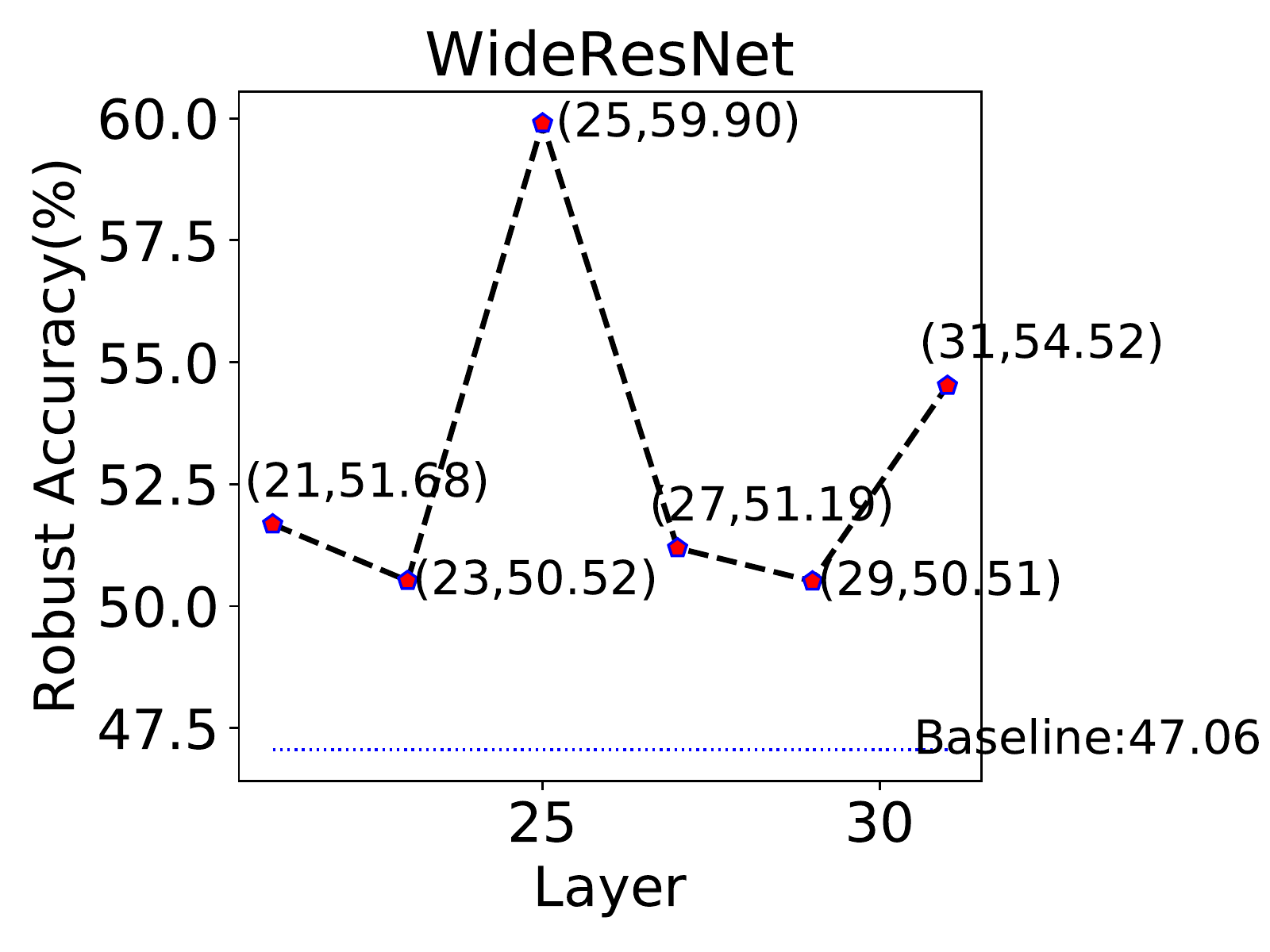}
                \end{minipage}
            }

\caption{The impact of EWAS position on CIFAR10.}
\label{fig:dflayer}
\end{figure}



\section{Conclusion}
\label{sec:con}

In this paper, we propose a new element-wise activation scaling (EWAS) method to improve CNNs' adversarial robustness. EWAS is a form of activation robustification techniques which can conduct a more fine-grained activation scaling. EWAS is a simple but very effective method to improve CNNs' robustness. It can be easily added to existing CNN models and be trained with the backbone network using an auxiliary loss function. The experimental results demonstrate that EWAS outperforms other two latest activation robustificiation techniques in terms of adverserial accuracy.

\bibliographystyle{named}
\bibliography{main}

\appendix
\newpage
\section{Algorithm of EWAS Training}
\label{apdx:algo}

\renewcommand{\algorithmicrequire}{\textbf{Input:}}
\renewcommand{\algorithmicensure}{\textbf{Output:}}

\begin{algorithm}
\caption{Adversarial training with EWAS}
\label{alg:weas}
\begin{algorithmic}[1]
\REQUIRE 
Dataset $S={(x_i,y_i)}_{i=1}^n$, CNN $F(\theta)$ with EWAS module, training epoch $T$
\ENSURE A robust CNN $F$\\
\FOR{$t={1,2,...,T}$}
\FOR{$(x_i,y_i)$ in $S$}
\STATE Generate adversarial example using PGD by solving inner-max problem in Eq. (\ref{eq:opt-all})
\STATE $\hat s=\text{ALC}({f^l}(x_i + \delta ),\theta^\text{ALC} )$
\STATE Generate $\bm{m}$ by Eq. (\ref{eq:CRS})
\STATE ${\tilde z^l} = {z^l} \otimes \bm{m}$
\STATE Give $\Tilde z^l$ to the next convolution, complete the forward-propagation and compute the overall loss 
\ENDFOR
\STATE Optimize all the parameter of model and EWAS by solving outer-min problem in Eq. (\ref{eq:opt-all}) using gradient descent
\ENDFOR
\end{algorithmic}
\end{algorithm}

\section{Loss function of EWAS-modified Model}
\label{apdx:lewas}

Here we show the loss functions of MART and TRADES, the two AT variants. As shown in Table \ref{tbl:loss}, $p$ denotes the prediction score of network F and $\hat p$ denotes the prediction score of EWAS module.

\begin{table*}[!bp]
\centering
\begin{tabular}{cc}
\Xhline{3\arrayrulewidth}
 Method & Loss function \\ \Xhline{3\arrayrulewidth}
\begin{tabular}[c]{@{}l@{}}AT\\ +EWAS\end{tabular}    & \begin{tabular}[c]{@{}c@{}}$L_{CE}(p(x+\delta,\theta),y)$\\$+\lambda \cdot L^{ALC}_{CE}(\hat p(x+\delta),y)$\end{tabular} \\ \hline
\begin{tabular}[c]{@{}l@{}}TRADES\\ +EWAS\end{tabular}       &\begin{tabular}[c]{@{}c@{}}$L_{CE}(p(x,\theta),y)+\beta \cdot L_{KL}(p(x,\theta),p(x+\delta,\theta))$\\$+ \lambda \cdot L^{ALC}_{CE}(\hat p(x),y) +\lambda \cdot \beta \cdot L^{ALC}_{KL}(\hat p(x),\hat p(x+\delta))$ \end{tabular}        \\ \hline
\begin{tabular}[c]{@{}l@{}}MART\\ +EWAS\end{tabular}     &  \begin{tabular}[c]{@{}c@{}}$L_{BCE}(p(x+\delta,\theta),y)+\beta \cdot L_{KL}(p(x,\theta),p(x+\delta,\theta))\cdot (1-p_y(x,\theta))$\\$+ \lambda \cdot L^{ALC}_{BCE}(\hat p(x+\delta),y) +\lambda \cdot \beta \cdot L^{ALC}_{KL}(\hat p(x),\hat p(x+\delta))\cdot (1-\hat p_y(x))$ \end{tabular}        \\ \Xhline{3\arrayrulewidth}
\end{tabular}
\caption{The loss function used for AT, TRADES, MART with EWAS module.}
\label{tbl:loss}
\end{table*}

\section{Details on Activation Visualizing}\label{apdx:DVAF}
We show the activation frequency and average activation magnitude.
\subsection{Activation Frequency}
We respectively performed natural and adversarial training on ResNet-18 under CIFAR-10 data set for 120 epochs with the SGD optimizer (momentum 0.9 and weight decay 0.0002). During adversarial training, We use adversarial data generated by PGD-10 attack ($\epsilon$ = 8/255, step size $\epsilon$/4, and random initialization). 

We use the output of the last residual block which also is the input of the global average pooling operation as the frequency visualization layer. The activation unit is valid if its activation magnitude is larger than 1\% of the maximum of all activation. For visualization, we select all samples of one class as the input samples, and the results are shown in descending order of channel frequencies of the natural samples.

\subsection{Activation Average Magnitude}

The training details follows those in activation frequency visualizing. We also use the output of the last residual block which also is the input of the global average pooling operation as the activation magnitude visualization layer. For a certain class, we calculate channel's max activation value for each samples and average it over all the same class samples. We also plot it in descending order of average magnitude of the natural samples.

\section{Experimental Setting Details}\label{apdx:exp}

The training setting of CIFS and CAS follows those in \cite{bai2021improving,cifs}, which are $\beta _{CIFS} = 2$ and $\beta _{CAS} = 2$.

\subsection{Experimental Details on CIFAR10}

We train models with 128 batch size using SGD optimizer (momentum 0.9 and weight decay 0.0002), and initial learning rate is 0.1. With different training methods, we set different training epoch and milestones with multiplicative factor of learning rate decay 0.1, as shown in Table \ref{tbl:ctd}. During AT, we set $\epsilon = 8/255$ and step size $\epsilon/4$ for PGD-10 to generate adversarial samples. For TRADES and MART, $\beta$ is 6.


\begin{table}[!htbp]
\centering
\begin{tabular}{ccc}
\Xhline{3\arrayrulewidth}
      & epochs & milestones \\ \Xhline{3\arrayrulewidth}
AT    & 120    & 60, 90     \\
TRADE & 85     & 75         \\
MART  & 90     & 60         \\ \Xhline{3\arrayrulewidth}
\end{tabular}
\caption{Training epochs and learning rate adjust milestones for CIFAR10 data set.}
\label{tbl:ctd}
\end{table}

\subsection{Experimental Details on SVHN}

 For SVHN dataset, we train model with 128 batch size using SGD optimizer (momentum 0.9 and weight decay 0.0005), and initial learning rate is 0.01, with different t raining method, we set same training epoch 120 and divided by 10 at 75-th and 90-th epoch. For training stage, we set $\epsilon = 8/255$ and step size $\epsilon/4$ for PGD-10 to generate adversarial samples. For TRADES and MART, $\beta$ is 6. 
 
 For SVHN evaluation, adversarial data are generated by FGSM, PGD-20 (20-steps PGD with random start), and C\&W (${L_\infty }$ version of C\&W optimized by PGD-30) , $\epsilon$ is $8/255$ and step size $\epsilon/10$.

\section{The Impact of EWAS Position}
\label{apdx:dflayer}

Here, we further show the experimental results of EWAS position evaluation. The natural and robust accuracies of ResNet-18 and WideResNet with different EWAS position against FGSM, PGD-20, C\&W (Table \ref{tbl:dflr} and Table \ref{tbl:dflw}). We train the model with $\lambda=0.01$, and the evaluation settings follow those in Appendix \ref{apdx:exp}.

\begin{table}[!htbp]
\centering
\begin{tabular}{ccccc}
\Xhline{3\arrayrulewidth}
Layer & Natural & FGSM   & PGD-20 & C\&W   \\ \Xhline{3\arrayrulewidth}
11    & 79.02  & 58.19 & 55.71 & 65.71 \\
13    & 83.58  & 62.73 & 58.77 & 72.77 \\
\underline{ 15 }    & 84.73   & 65.78  & 64.84  & 82.35  \\
17    & 83.73   & 62.67  & 56.63  & 71.31 \\ \Xhline{3\arrayrulewidth}
\end{tabular}
\caption{Robustness comparison of the EWAS module after different layers of ResNet-18 on CIFAR10. We reported the robust accuracy (\%) at the last epoch. The final selected layer is marked with underline.}
\label{tbl:dflr}
\end{table}

\begin{table}[!htbp]
\centering
\begin{tabular}{ccccc}
\Xhline{3\arrayrulewidth}
Layer & Natural & FGSM   & PGD-20 & C\&W   \\ \Xhline{3\arrayrulewidth}
21    & 85.63  & 61.71 & 51.68 & 58.00   \\
23    & 85.70   & 62.46 & 50.52 & 55.43 \\
\underline{ 25 }    & 87.12  & 64.05 & 59.90  & 73.01 \\
27    & 86.66  & 62.74 & 51.19 & 60.68 \\
29    & 86.36  & 61.68 & 50.51 & 58.62 \\
31    & 85.90   & 65.06  & 54.52  & 62.60 \\ \Xhline{3\arrayrulewidth}
\end{tabular}
\caption{Robustness comparison of the EWAS module after different layers of WideResNet on CIFAR10. The final selected layer is marked with underline.}
\label{tbl:dflw}
\end{table}

\section{The Performance of EWAS}
\label{apdx:pewas}

\begin{figure*}[!htbp]
        \centering
            \subfigure[Before EWAS]
            {
                \begin{minipage}[b]{0.2\textwidth}
                \includegraphics[width=1\textwidth]{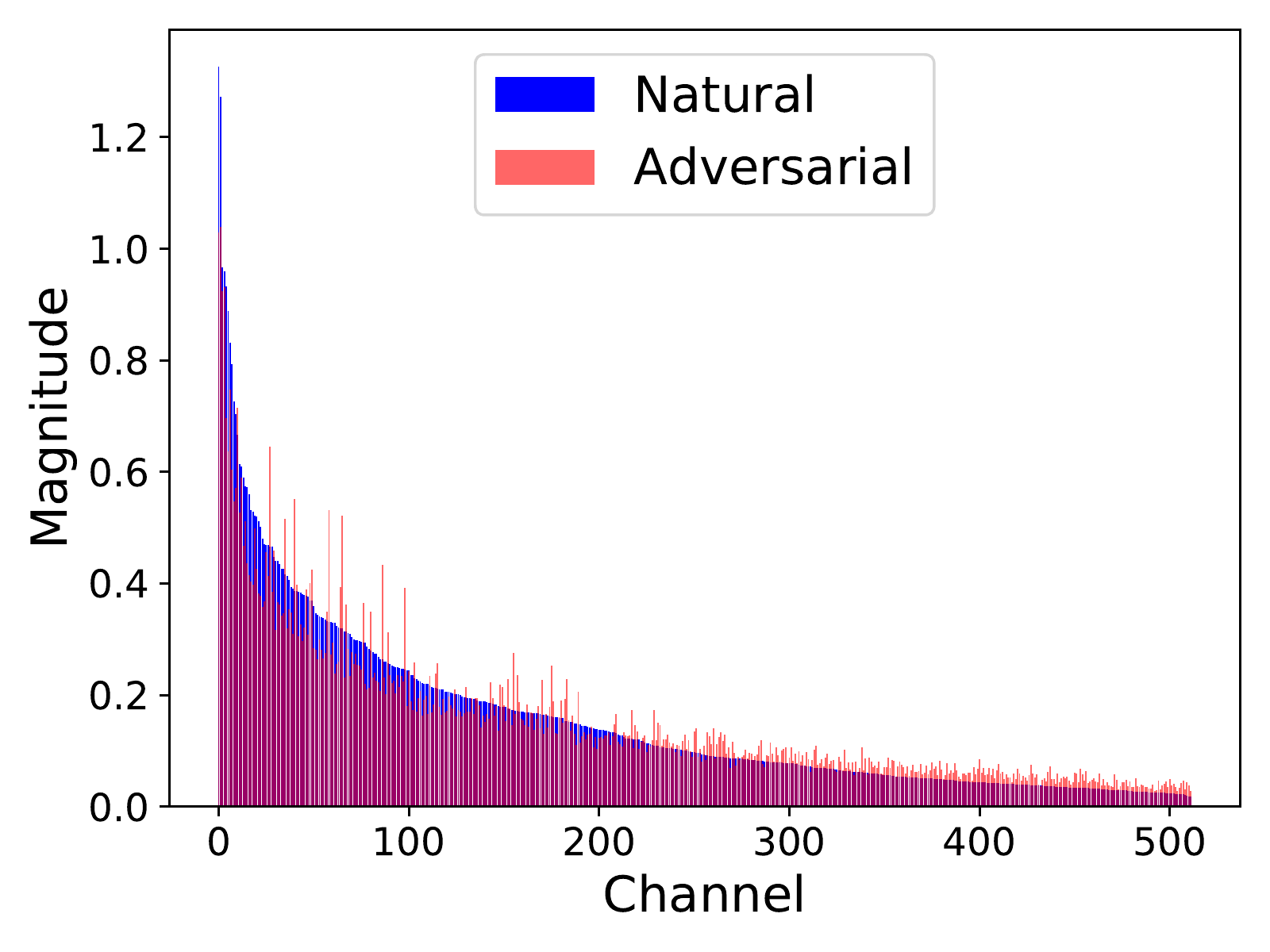}
                \includegraphics[width=1\textwidth]{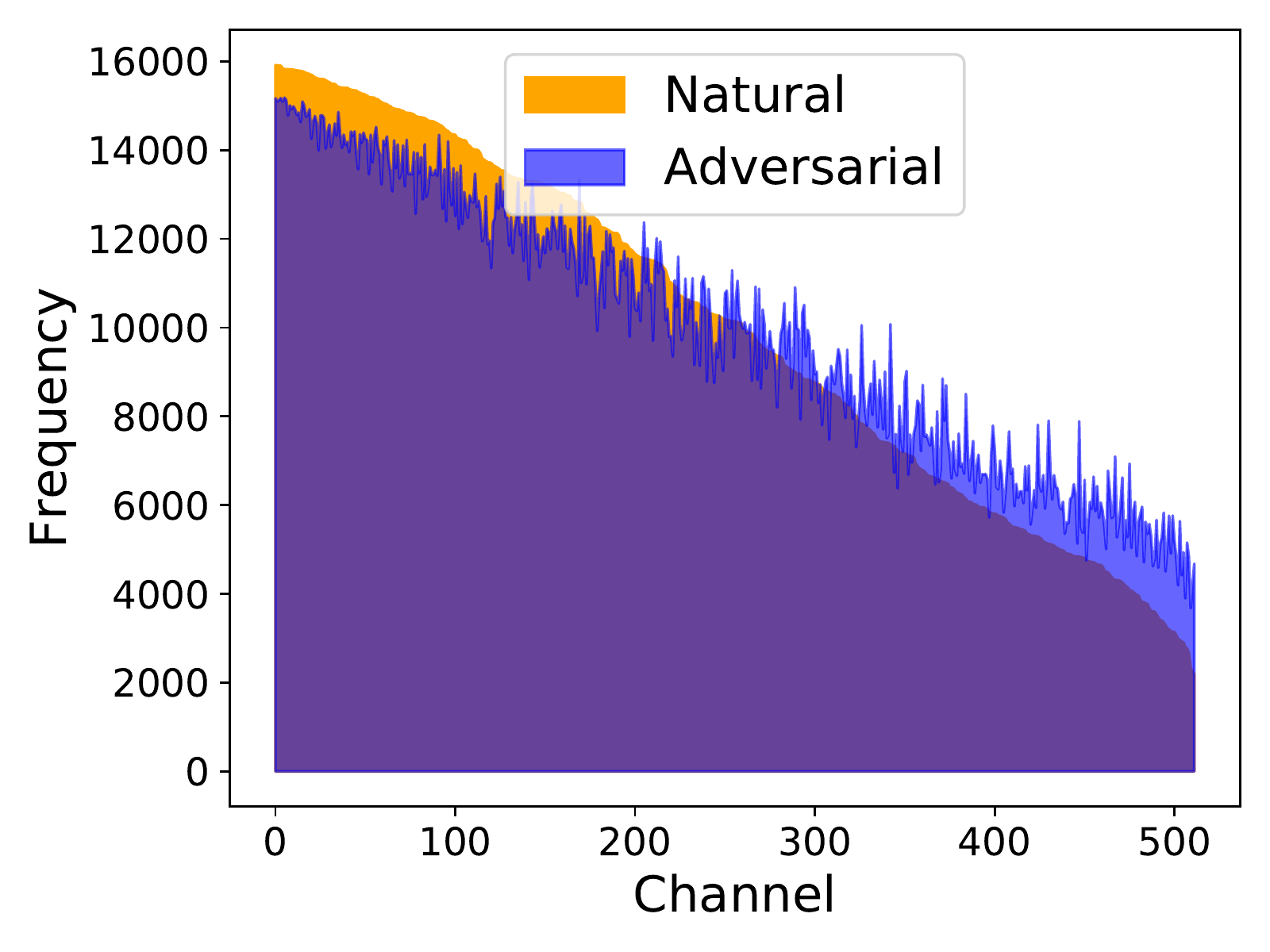}
                \end{minipage}
            }
            \subfigure[After EWAS] 
            {
                \begin{minipage}[b]{0.2\textwidth}
                \includegraphics[width=1\textwidth]{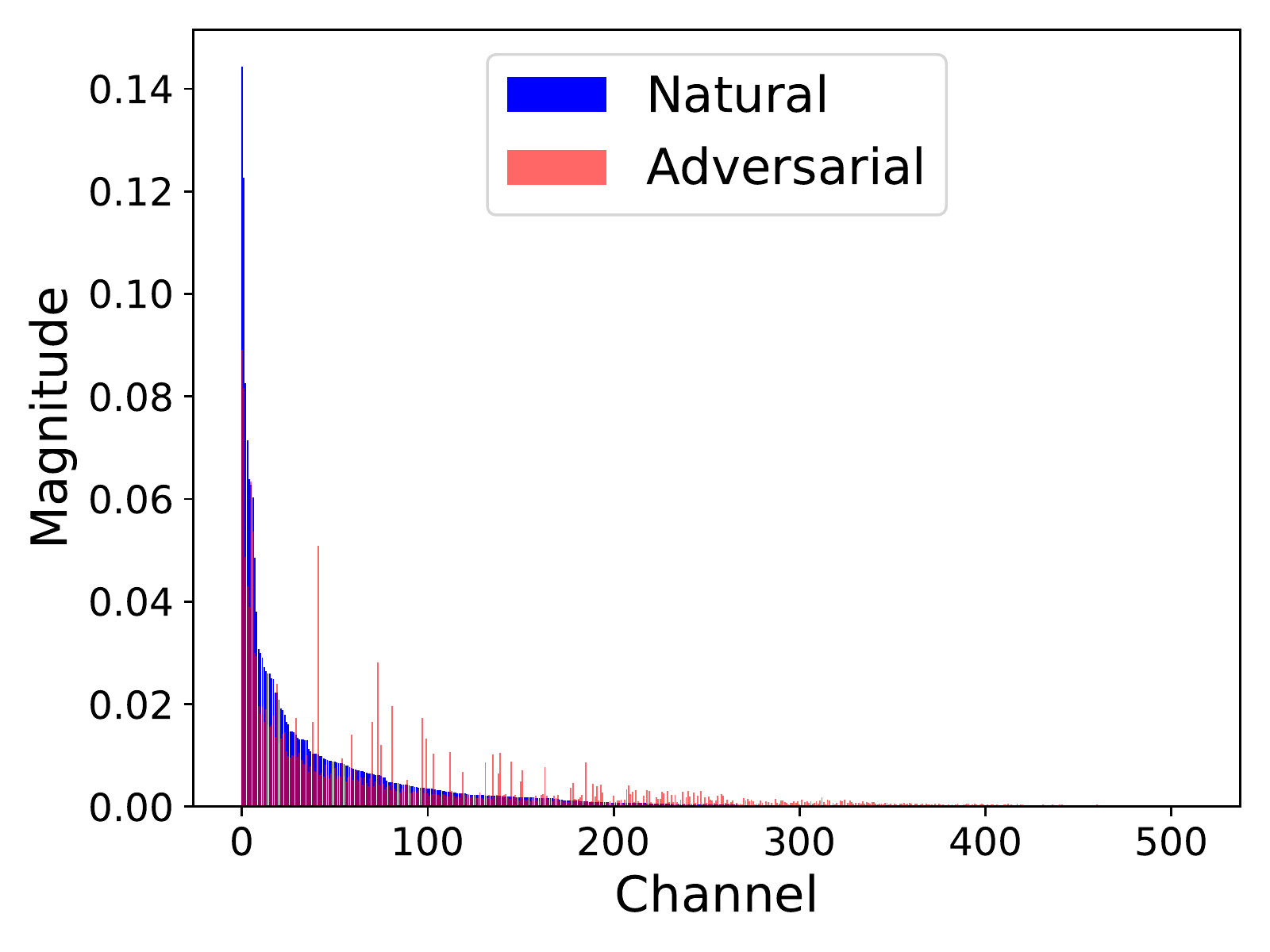}
                \includegraphics[width=1\textwidth]{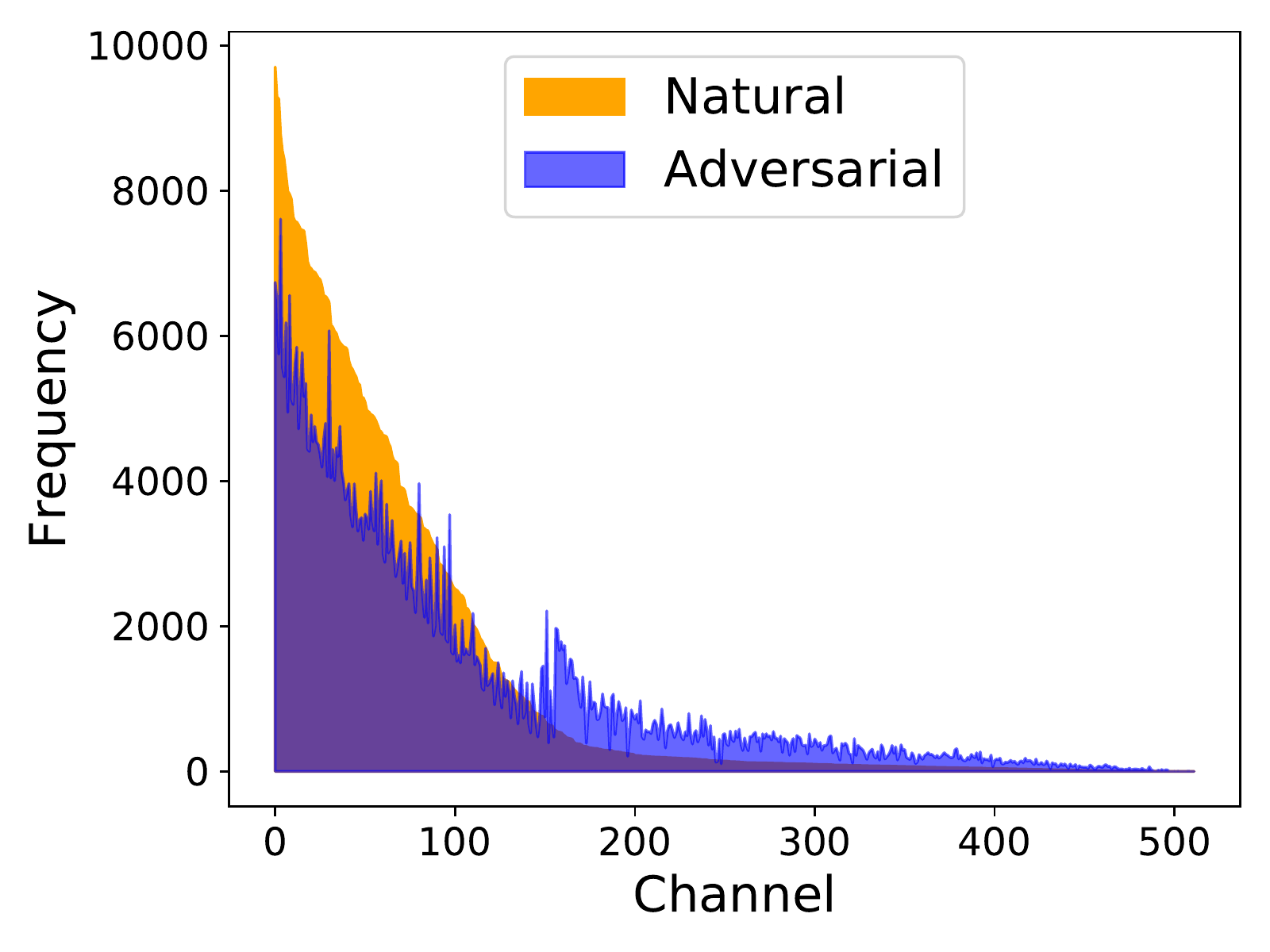}
                \end{minipage}
            }
            \subfigure[The penultimate] 
            {
                \begin{minipage}[b]{0.2\textwidth}
                \includegraphics[width=1\textwidth]{figures/my_at_last_mag.pdf}
                \includegraphics[width=1\textwidth]{figures/myat_channel_class_0.pdf}
                \end{minipage}
            }
\caption{Comparison of activation average magnitude and frequency between adversarial and natural examples before and after EWAS scaling and the penultimate layer. Natural samples are from CIFAR-10 "airplane" class.}
\label{fig:bf}
\end{figure*}

To visualize how the EWAS module affects the intermediate activation, we visualize the  average activation magnitude and frequency of ResNet-18 before and after EWAS scaling on CIFAR10 under AT. As shown in Fig \ref{fig:bf}, after EWAS scaling, both the magnitude and frequency have dropped drastically. From the figure, we can see that before EWAS, the activation magnitude is high, and after EWAS the activation magnitude is suppressed. Along the forward propagation, the activation of the penultimate layer shows different distributions between natural examples and adversaries.

 \section{Attack Impact of $\lambda$}
 \label{apdx:evlam}

\begin{table}[!htbp]
\centering
\begin{tabular}{ccccc}
\Xhline{3\arrayrulewidth}
$\lambda$     & Natural & FGSM   & PGD-20 & C\&W   \\ \Xhline{3\arrayrulewidth}
0       & 84.73  & 86.09 & 85.33 & 84.60  \\
\underline{ 0.01 }   &         & 65.78  & 64.84  & 82.35  \\
0.1     &         & 63.29 & 56.22 & 60.91 \\
0.5     &         & 62.71 & 47.55 & 47.90  \\
1       &         & 62.71 & 46.99 & 46.95 \\
2       &         & 62.71 & 46.78 & 46.91 \\
3       &         & 62.71 & 46.72 & 46.87 \\
5       &         & 62.71 & 46.69 & 46.87 \\
10      &         & 62.71 & 46.66 & 46.79 \\
Vanilla & 84.47   & 61.09  & 44.33  & 44.70  \\ \Xhline{3\arrayrulewidth}
\end{tabular}
\caption{Robustness comparison of the different evaluation $\lambda$ of ResNet-18 on CIFAR10. We reported the robust accuracy (\%) at the last epoch. The training $\lambda$ is marked with underline.}
\label{tbl:dfelr}
\end{table}
 
\begin{table}[!htbp]
\centering
\begin{tabular}{ccccc}
\Xhline{3\arrayrulewidth}
$\lambda$     & Natural                   & FGSM                      & PGD-20                    & C\&W                      \\ \Xhline{3\arrayrulewidth}
0       & 87.12                    & 83.96                    & 83.61                    & 83.66                    \\
\underline{ 0.01 }   &                      & 64.05                     & 59.90                      & 73.01                     \\
0.1     &                           & 63.50                    & 48.88                    & 50.42                    \\
0.5     &                           & 63.49                    & 47.27                    & 48.23                    \\
1       &                           & 63.50                     & 47.21                    & 48.08                    \\
2       &                           & 63.50                     & 47.20                     & 48.09                    \\
3       &                           & 63.50                     & 47.19                    & 48.07                    \\
5       &                           & 63.50                     & 47.20                     & 48.06                    \\
10      &                           & 63.50                     & 47.19                    & 48.05                    \\
Vanilla & 86.65 & 63.71 & 47.06  & 45.75 \\ \Xhline{3\arrayrulewidth}
\end{tabular}
\caption{Robustness comparison of the different evaluation $\lambda$ of WideResNet on CIFAR10. The training $\lambda$ is marked with underline.}
\label{tbl:dfelw}
\end{table}

We set different $\lambda=[0,0.01,0.1,0.5,1,2,3,5,10]$ to control the attack degree on the EWAS, where the larger the $\lambda$, the stronger the attack effect on the EWAS module. In other words, as the $\lambda$ increases, the attack will focus on the EWAS module until the EWAS module is compromised, which means the model can only rely on its own robustness. 

Here, we report the natural and robust accuracies of EWAS-modified ResNet-18 and WideResNet against FGSM, PGD-20, C\&W (Table \ref{tbl:dfelr} and Table \ref{tbl:dfelw}). 
When the adversary only takes the backbone classification loss as the maximization goal ($\lambda = 0$), it is very likely that the attack will fail. As the attack focuses on the EWAS loss, the robustness of the model will gradually decrease, but its robustness is still higher than the vanilla. We can see that EWAS plays an important role in the robustness of the model.

\section{The Visualization of WideResNet}
\label{apdx:vw}

\begin{figure}[!htbp]
        \centering
            \subfigure[AT] 
            {
                \begin{minipage}[b]{0.2\textwidth}
                \includegraphics[width=1\textwidth]{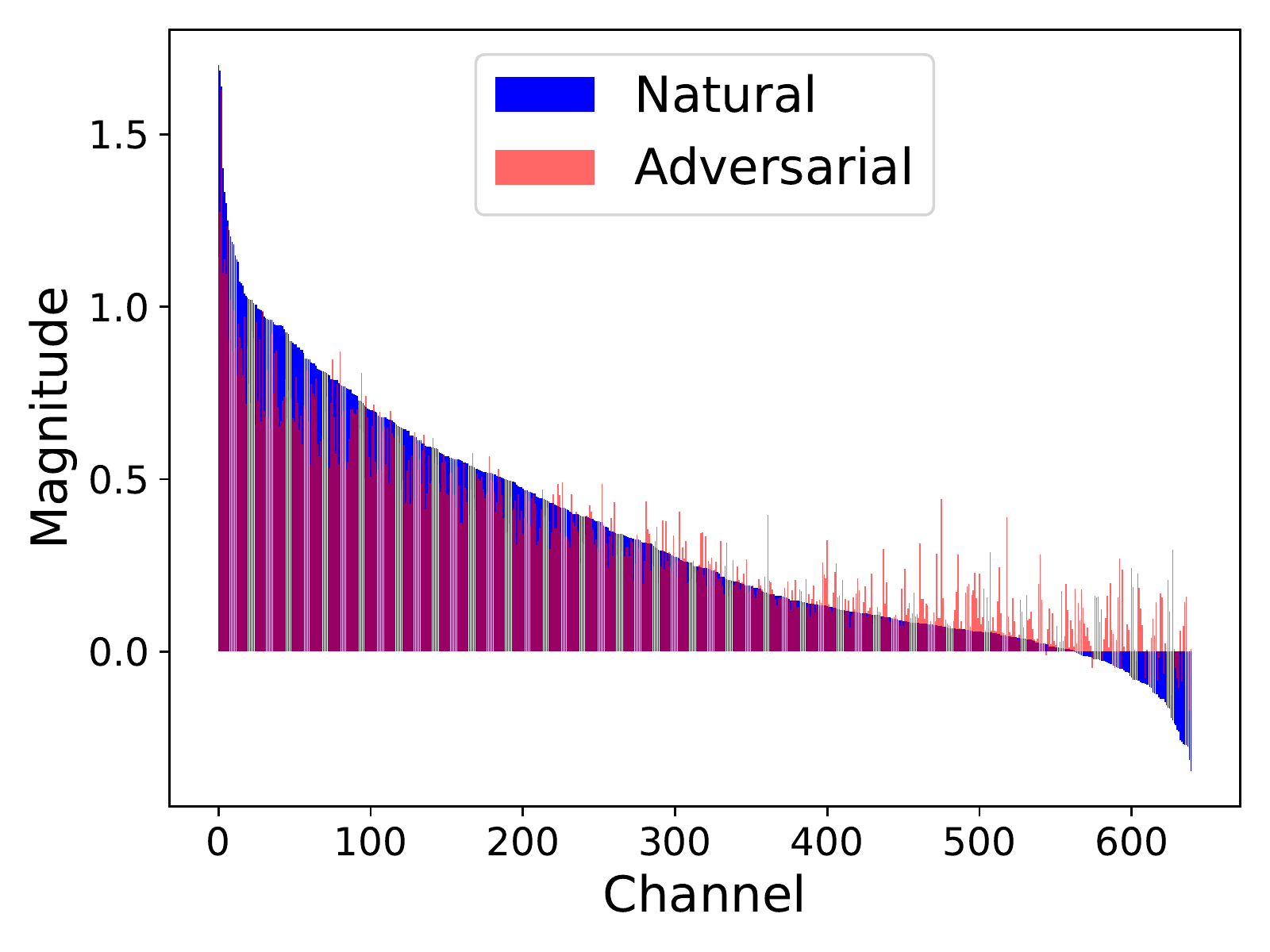}
                \includegraphics[width=1\textwidth]{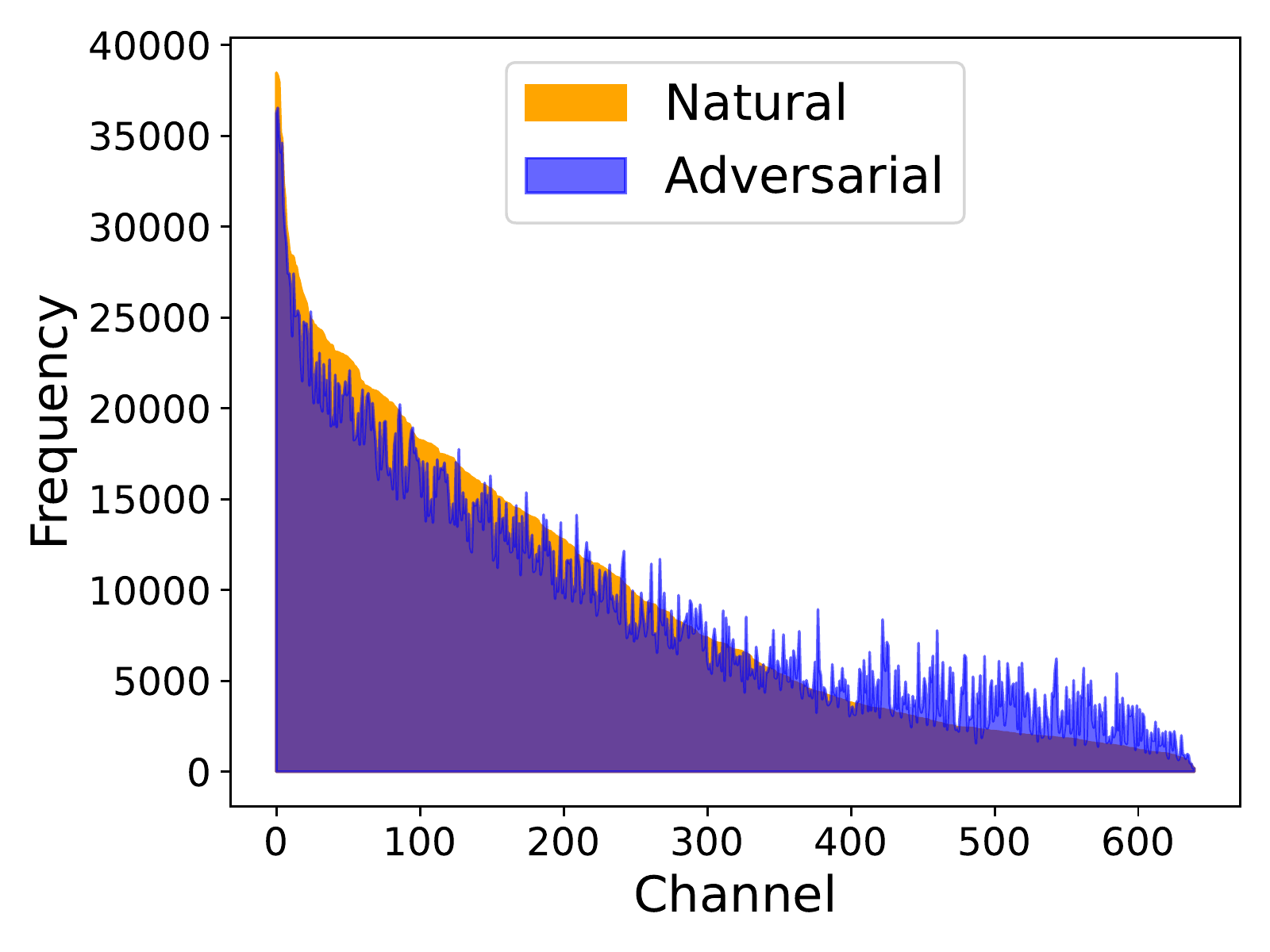}
                \end{minipage}
            }
           \subfigure[EWAS] 
            {
                \begin{minipage}[b]{0.2\textwidth}
                \includegraphics[width=1\textwidth]{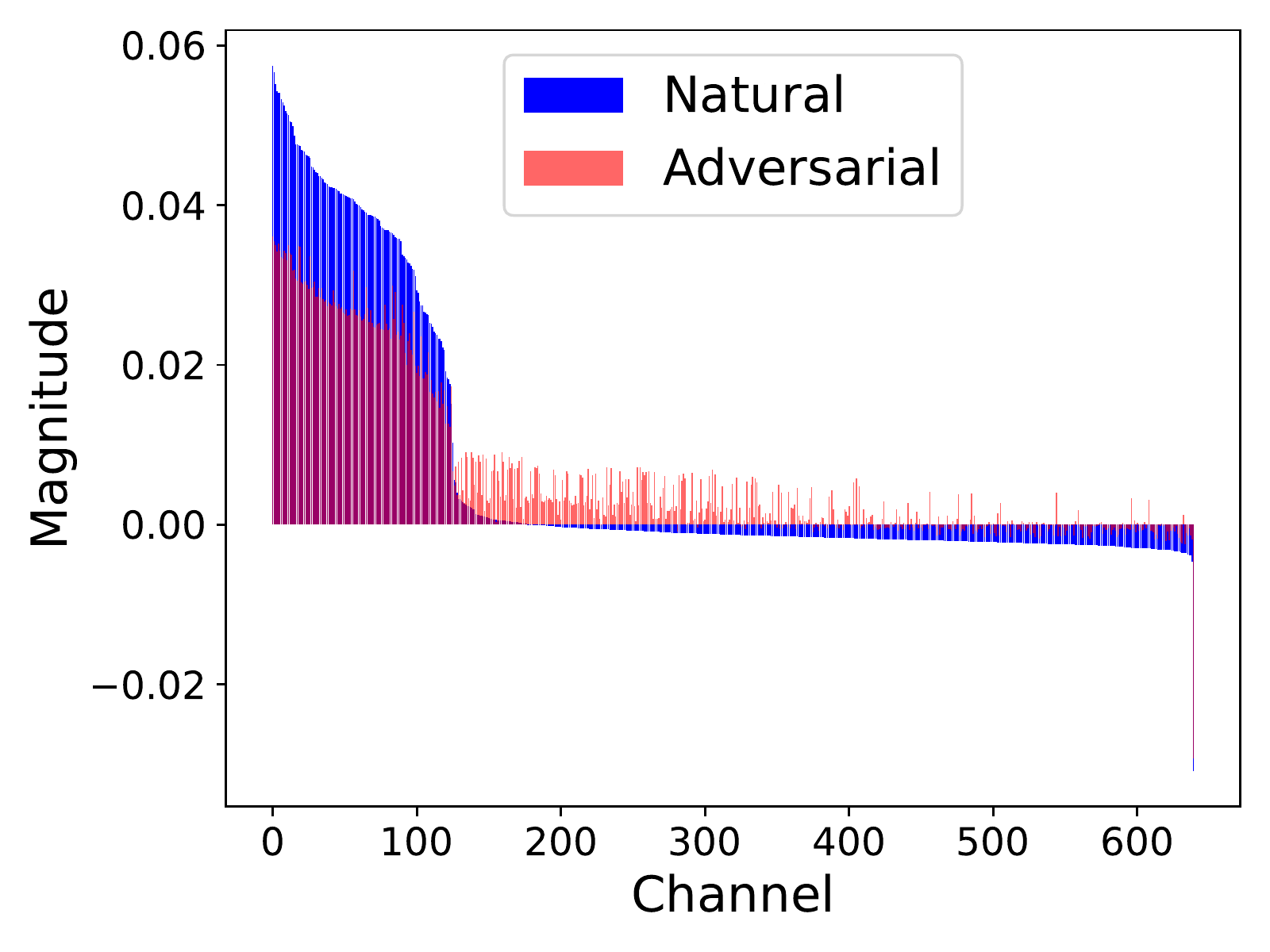}
                \includegraphics[width=1\textwidth]{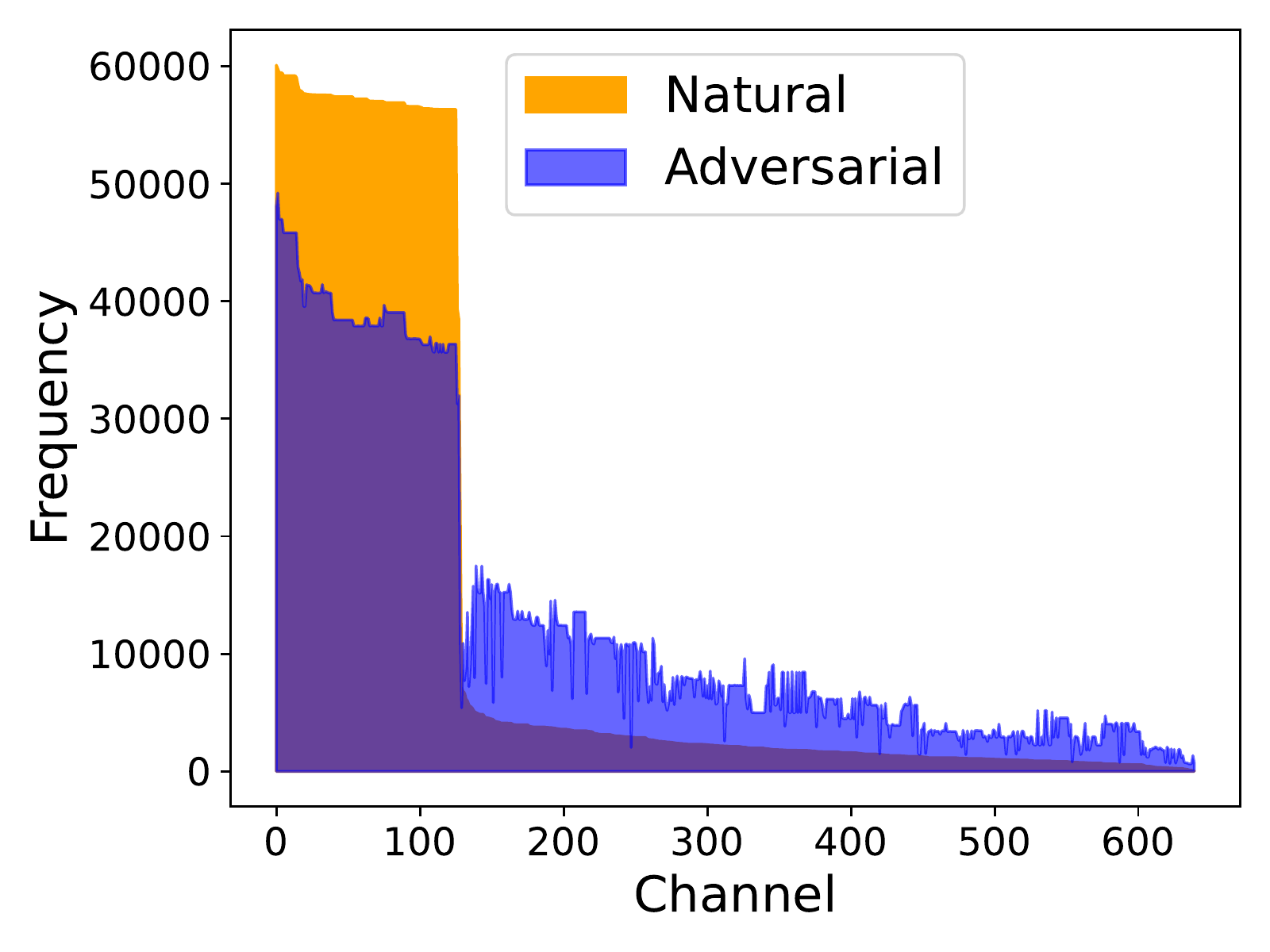}
                \end{minipage}
            }

\caption{Comparison of activation magnitude and frequency between adversarial and natural samples on different defense methods on WideResNet. Natural samples are from CIFAR-10 "airplane" class.}
\label{fig:compare-w}
\end{figure}

Here we visualize the activation magnitude and frequency of the penultimate layer of WideResNet, as shown in Fig \ref{fig:compare-w}. The results also show EWAS presents more difference between natural examples and adversarial examples.

\end{document}